\documentclass{article}

\usepackage{arxiv}

\usepackage[utf8]{inputenc} 
\usepackage[T1]{fontenc}    
\usepackage{hyperref}       
\usepackage{url}            
\usepackage{booktabs}       
\usepackage{amsfonts}       
\usepackage{nicefrac}       
\usepackage{microtype}      
\usepackage{lipsum}
\usepackage{amsmath}
\usepackage{amsthm}
\usepackage{amssymb}
\usepackage{bm}
\usepackage{caption}
\usepackage{subcaption}
\usepackage{graphicx}
\usepackage[ruled,vlined]{algorithm2e}
\include{pythonlisting}
\usepackage{xcolor}
\usepackage{mathtools}

\title{Understanding and mitigating gradient pathologies in physics-informed neural networks}

\author{
  Sifan Wang \\
  Graduate Group in Applied Mathematics \\
  and Computational Science \\
  University of Pennsylvania\\
  Philadelphia, PA 19104 \\
  \texttt{sifanw@sas.upenn.edu} \\
     \And
  Yujun Teng \\
  Department of Mechanichal Engineering \\
  and Applied Mechanics\\
  University of Pennsylvania\\
  Philadelphia, PA 19104 \\
  \texttt{yjteng@seas.upenn.edu } \\
   \And
  Paris Perdikaris \\
  Department of Mechanichal Engineering \\
  and Applied Mechanics\\
  University of Pennsylvania\\
  Philadelphia, PA 19104 \\
  \texttt{pgp@seas.upenn.edu} \\
}

\begin{document}
\maketitle

\begin{abstract}
The widespread use of neural networks across different scientific domains often involves constraining them to satisfy certain symmetries, conservation laws, or other domain knowledge. Such constraints are often imposed as soft penalties during model training and effectively act as domain-specific regularizers of the empirical risk loss. Physics-informed neural networks is an example of this philosophy in which the outputs of deep neural networks are constrained to approximately satisfy a given set of partial differential equations. In this work we review recent advances in scientific machine learning with a specific focus on the effectiveness of physics-informed neural networks in predicting outcomes of physical systems and discovering hidden physics from noisy data. We will also identify and analyze a fundamental mode of failure of such approaches that is related to numerical stiffness leading to unbalanced back-propagated gradients during model training. To address this limitation we present a learning rate annealing algorithm that utilizes gradient statistics during model training to balance the interplay between different terms in composite loss functions. We also propose a novel neural network architecture that is more resilient to such gradient pathologies. Taken together, our developments provide new insights into the training of constrained neural networks and consistently improve the predictive accuracy of physics-informed neural networks by a factor of 50-100x across a range of problems in computational physics. All code and data accompanying this manuscript are publicly available at \url{https://github.com/PredictiveIntelligenceLab/GradientPathologiesPINNs}.
\end{abstract}


\section{Introduction}
\label{sec:intro}
Thanks to breakthrough results across a diverse range of scientific disciplines \cite{krizhevsky2012imagenet,cho2014learning,alipanahi2015predicting,baldi2014searching,silver2017mastering}, deep learning 
is currently influencing the way we process data, recognize patterns, and build predictive models 
of complex systems. Many of these predictive tasks are currently being tackled using over-parameterized, black-box discriminative models such as deep neural networks, in which interpretability and robustness is often sacrificed in favor of flexibility in representation and scalability in computation. Such models have yielded remarkable results in data-rich domains \cite{krizhevsky2012imagenet, donahue2015long, radford2015unsupervised}, yet their effectiveness in the small data regime still remains questionable, motivating the question of how can one endow these powerful black-box function approximators with prior knowledge and appropriate inductive biases. 

Attempts to address this question are currently defining two distinct schools of thought. The first pertains to efforts focused on designing specialized neural network architectures that implicitly embed any prior knowledge and inductive biases associated with a given predictive task. Without a doubt, the most celebrated example in this category is convolutional neural networks \cite{lecun1995convolutional,goodfellow2016deep} which have revolutionized the field of computer vision by craftily respecting invariance along the groups of symmetries and distributed pattern representations found in natural images \cite{mallat2016understanding}. Another example includes covariant neural networks \cite{kondor2018covariant}, that are tailored to conform with the rotation and translation invariance present in many-body molecular systems.  Despite their remarkable effectiveness, such approaches are currently limited to tasks that are characterized by relatively simple and well-defined symmetry groups, and often require craftsmanship and elaborate implementations. Moreover, their extension to more complex tasks is challenging as the underlying  invariance that characterize many physical systems (e.g., fluid flows) are often poorly understood or hard to implicitly encode in a neural architecture. 

The second school of thought approaches the problem of endowing a neural net with prior knowledge from a different angle. Instead of designing specialized architectures that implicitly bake in this knowledge, current efforts aim to impose such constraints in a soft manner by appropriately penalizing the loss function of conventional neural network approximations denoted by $f_\theta(\cdot)$ and typically parametrized by a set of weights and biases $\theta$. These penalty constraints lead to loss functions taking the general form

\begin{equation}
\label{eq:soft_penalty}
\mathcal{L}(\theta) := \frac{1}{N_u}\sum\limits_{i=1}^{N_u} [\bm{u}_i-f_{\theta}(\bm{x}_i)]^2 + \frac{1}{\lambda}\mathcal{R}[f_{\theta}(\bm{x})],
\end{equation}
where the empirical risk loss over a given set of input-output pairs $\{\bm{x}_i, \bm{u}_i\}$, $i=1,\dots,N_u$ is penalized by some appropriate functional $\mathcal{R}[f_{\theta}(\bm{x})]$ that is designed to constraint the outputs of the neural network to satisfy a set of specified conditions, as controlled by the regularization parameter $\lambda$. A representative example of this approach is {\em physics-informed neural networks} \cite{raissi2019physics,yang2019adversarial,kharazmi2019variational,tartakovsky2018learning}, in which the outputs of a neural network are constrained to approximately satisfy a system of partial differential equations (PDE) by using a regularization functional $\mathcal{R}[f_{\theta}(\bm{x})]$ that typically corresponds to the residual or the variational energy of the PDE system under the neural network representation. This framework has enabled the solution of forward and inverse problem in computational physics by reviving the original ideas in \cite{psichogios1992hybrid,lagaris1998artificial} using modern software developments in reverse-mode differentiation \cite{baydin2018automatic} in order to automatically compute any derivatives present in the PDE operator. 
Although such approaches appear seemingly straightforward and have yielded remarkable results across a range of problems in computational science and engineering \cite{raissi2018hidden, raissi2018deep, zhu2019physics, geneva2020modeling, sirignano2018dgm, tripathy2018deep, kissas2020machine, sun2019surrogate} , the effects of the regularization mechanism in equation \ref{eq:soft_penalty} remain poorly understood, and in several cases can even lead to unstable and erroneous predictions (for e.g. see remarks in \cite{raissi2018deep, sun2019surrogate}).

In this work, we use physics-informed neural networks as a test-bed for analyzing the performance of constrained neural networks trained using regularized loss functions in the form of equation \ref{eq:soft_penalty}. Our specific contributions can be summarized in the following points:

\begin{itemize}
    \item Our analysis reveals a fundamental mode of failure in physics-informed neural networks related to stiffness in the gradient flow dynamics.
    \item This leads to an unstable imbalance in the magnitude of the back-propagated gradients during model training using gradient descent.
    \item We propose a simple solution based on an adaptive learning rate annealing algorithm that aims to balance the interplay between data-fit and regularization.
    \item We also propose a  novel neural network architecture that has less stiffness than the convention fully-connected neural network.
    \item We systematically test the proposed ideas and demonstrate consistent improvements in the predictive accuracy of physics-informed neural networks by a factor of 50-100x across a range of problems in computational physics.
\end{itemize}
Taken all together, our developments provide new insight into the training of constrained neural networks that can help us endow deep learning tools with prior knowledge and reduce the energy barrier for adapting them to new scientific domains. 

The paper is structured as follows. In section \ref{sec:methods}, we first present a brief overview of the physics-informed neural networks (PINNs) following the original formulation of Raissi {\it et. al.} \cite{raissi2019physics}. Next, we introduce a simple benchmark problem that can guide our analysis and highlight the difficulties introduced by stiffness in the gradient flow dynamics of physics-informed neural networks, see sections \ref{sec:gradient_pathologies} - \ref{sec:stiffness}. To address these difficulties we proposed an adaptive learning rate algorithm along with a novel fully-connected neural architecture, as described in sections \ref{sec:adaptive} and \ref{sec:improved_FC}. In section \ref{sec:results} we present the detailed evaluation of our proposed algorithm and neural network architecture across a range of representative benchmark examples. Finally, in section \ref{sec:discussion}, we summarize our findings and provide a discussion on potential pitfalls and promising future directions. All code and data accompanying this manuscript are publicly available at \url{https://github.com/PredictiveIntelligenceLab/GradientPathologiesPINNs}.

\section{Methods}
\label{sec:methods}

\subsection{A primer in physics-informed neural networks}

Physics-informed neural networks (PINNs) \cite{raissi2019physics} aim at inferring a continuous latent function $\bm{u}(\bm{x},t)$ that arises as the solution to a  system of nonlinear partial differential equations (PDE) of the general form
\begin{align}
\label{eq:PDE}
    \begin{split}
     &\bm{u}_t + \mathcal{N}_{\bm{x}}[\bm{u}] = 0, \ \  \bm{x} \in \Omega, t \in [0, T] \\
     &\bm{u}(\bm{x}, 0) = h(\bm{x}), \ \ x \in \Omega \\
     &\bm{u}(\bm{x}, t) = g(\bm{x}, t), \ \ t \in [0, T], \ \  \bm{x} \in \partial \Omega
    \end{split}
\end{align}
where $\bm{x}$ and $t$ denote space and time coordinates, subscripts denote partial differentiation, $\mathcal{N}_{\bm{x}}[\cdot]$ is a nonlinear differential operator, $\Omega$ is a subset of $\mathbb{R}^D$, and $\partial \Omega$ is the boundary of $\Omega$. Following the original work of \cite{raissi2019physics}, we then proceed by approximating $\bm{u}(\bm{x},t)$ by a deep neural network $f_{\theta}(\bm{x},t)$, and define the residual of equation  \ref{eq:PDE} as
\begin{align}
\label{eq:residual}
    \bm{r}_{\theta}(\bm{x}, t) := \frac{\partial}{\partial t}f_{\theta}(\bm{x},t) + \mathcal{N}_{\bm{x}}[f_{\theta}(\bm{x},t)],
\end{align}
where the partial derivatives of the neural network representation with respect to the space and time coordinates can be readily computed to machine precision using reverse mode differentiation \cite{baydin2018automatic}. Notice how the neural network parameters $\theta$ (i.e. the weights and biases of the neural network) are shared between the representation of the latent solution $\bm{u}(\bm{x},t)$, and the PDE residual $\bm{r}(\bm{x},t)$. A good set of candidate parameters can be identified via gradient descent using a composite loss function of the general form 

\begin{align}
\label{eq:loss}
    \mathcal{L}(\theta) :=  \mathcal{L}_r(\theta) + \sum\limits_{i=1}^{M}\lambda_i \mathcal{L}_i(\theta), 
\end{align} 
where $\mathcal{L}_r(\theta)$ is a loss term that penalizes the PDE residual, and $\mathcal{L}_i(\theta)$, $i=1,\dots,M$ correspond to data-fit terms (e.g., measurements, initial or boundary conditions, etc.). For a typical initial and boundary value problem, these loss functions would take the specific form
\begin{align}
\label{eq: loss_r}
    &\mathcal{L}_r = \frac{1}{N_r} \sum_{i=1}^{N_r} [\bm{r}(\bm{x}_r^i, t_r^i)]^2 \\
    \label{eq: loss_ub}
     &\mathcal{L}_{u_b} =\frac{1}{N_b} \sum_{i=1}^{N_b}[\bm{u}(\bm{x}_b^i, t_b^i) - g_b^i]^2, \\
     \label{eq: loss_u0}
    &\mathcal{L}_{u_0} = \frac{1}{N_0} \sum_{i=1}^{N_0}[\bm{u}(\bm{x}_0^i, 0) - h_0^i)]^2, 
\end{align}
where $\{\bm{x}_0^i, h_0^i) \}_{i=1}^{N_0}$ denotes the initial data, $\{(\bm{x}_b^i, t_b^i), g_b^i\}_{i=1}^{N_b}$ denotes the boundary data, and $\{(\bm{x}_r^i, t_r^i), \bm{0}\}_{i=1}^{N_r}$ a set of collocation points that are randomly placed inside the domain $\Omega$ in order to minimize the PDE residual. Consequently, $\mathcal{L}_r$ penalizes the equation not being satisfied on the collocation points. Moreover, $\mathcal{L}_{u_b}$  and $\mathcal{L}_{u_0} $ enforces the boundary conditions and the initial conditions respectively. As $\mathcal{L}(\theta)$ is typically minimized used stochastic gradient descent, an very large number of training points ($\mathcal{O}(10^5-10^8$) can be sampled as the locations of $\bm{x}_0^i, (\bm{x}_b^i, t_b^i), (\bm{x}_r^i, t_r^i)$ can be randomized within each gradient descent iteration. The ultimate goal of this procedure it to construct a neural network representation $f_{\theta}(\bm{x},t)$ for which $\mathcal{L}(\theta)$ is as close to zero as possible.

\subsection{Gradient pathologies in physics-informed neural networks}
\label{sec:gradient_pathologies}

Despite a series of promising results \cite{raissi2018hidden, raissi2018deep, sirignano2018dgm, tripathy2018deep, kissas2020machine, sun2019surrogate, yang2019adversarial}, the original formulation of Raissi {\it et. al.} \cite{raissi2019physics} often has difficulties in constructing an accurate approximation to the exact latent solution $\bm{u}(\bm{x},t)$ for reasons that yet remain poorly understood. These pathologies can arise even in the simplest possible setting corresponding to solving classical linear elliptic equations. As an example, let us consider the Helmholtz equation in two space dimensions
\begin{align}
    \label{eq:Helmholtz}
    & \Delta u(x,y) + k^2 u(x,y) = q(x,y), \ \ (x,y) \in \Omega:= (-1, 1) \\
    &u(x,y) = h(x,y), \ \  (x,y)\in \partial \Omega 
\end{align}
where $\Delta$ is the Laplace operator. One can easily fabricate an exact solution to this problem taking the form  $u(x, y) =\sin(a_1 \pi x) \sin(a_2 \pi y)$,
corresponding to a source term of the form 
\begin{align}
    q(x,y) = - (a_1 \pi)^2 \sin(a_1 \pi x) \sin(a_2 \pi y) - (a_2 \pi)^2 \sin(a_1\pi x) \sin(a_2 \pi y) + k^2 \sin(a_1\pi x) \sin(a_2 \pi y)
\end{align}
 where we take $a_1 =1$ and $a_2 =4$.
A PINN approximation to solving equation \ref{eq:Helmholtz} can be constructed by parametrizing its solution with a deep neural network $f_{\theta}(x,y)$, whose parameters $\theta$ can be identified by minimizing a composite loss function that aims to fit the known boundary conditions, while also penalizing the Helmholtz equation residual inside the domain $\Omega$, i.e.,
\begin{align}
\label{eq:loss_Helmholtz}
         \mathcal{L}(\theta) &=   \mathcal{L}_r(\theta) +  \mathcal{L}_{u_b}(\theta) 
\end{align}

\begin{figure}
    \centering
    \includegraphics[width = \textwidth]{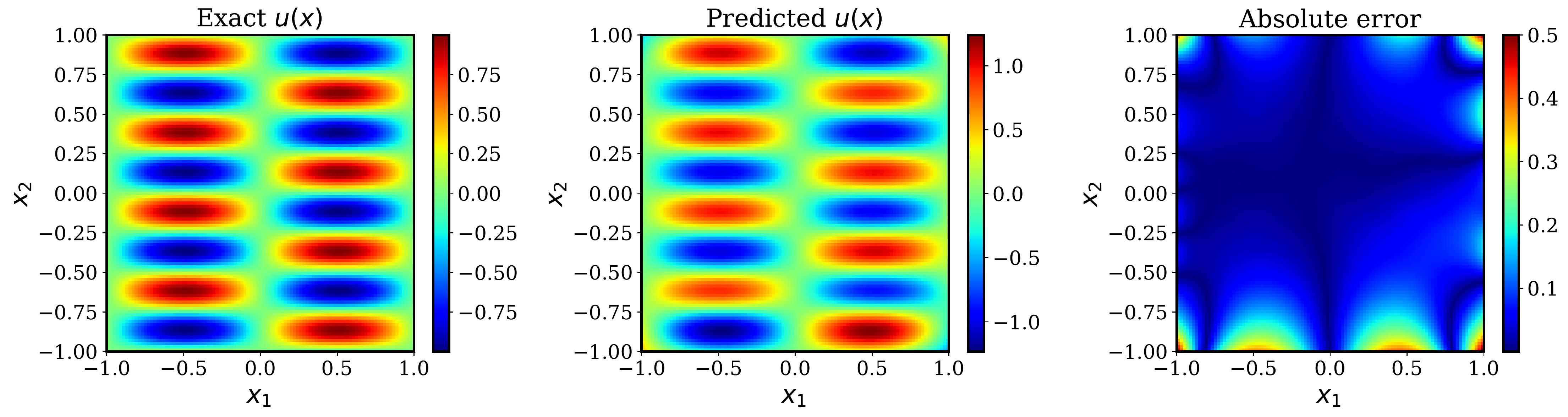}
    \caption{{\em Helmholtz equation:} Exact solution versus the prediction of a conventional physics-informed neural network model with 4 hidden layers and 50 neurons each layer after 40,000 iterations of training with gradient descent (relative $L^2$-error: 1.81e-01).}
    \label{fig:Helmholtz_prediction}
\end{figure}

Without loss of generality, let us consider an example prediction scenario in which $f_{\theta}(x,y)$ is a 4-layer deep fully connected neural network with 50 neurons per layer and a hyperbolic tangent activation function. We train this network for $40,000$ stochastic gradient descent steps by minimizing the loss of equation \ref{eq:loss_Helmholtz} using the Adam optimizer \cite{kingma2014adam} with an initial learning rate of $10^{-3}$ and a decreasing annealing schedule. In figure \ref{fig:Helmholtz_prediction} we compare the predictions of this trained model against the exact solution for this problem, and report the point-wise absolute discrepancy between the two. It is evident that the PINN approximation does a poor job at fitting the boundary conditions, leading to a $15.7\%$ prediction error measured in the relative $L^2$ norm.

\begin{figure}
    \centering
    \includegraphics[width = 0.8 \textwidth]{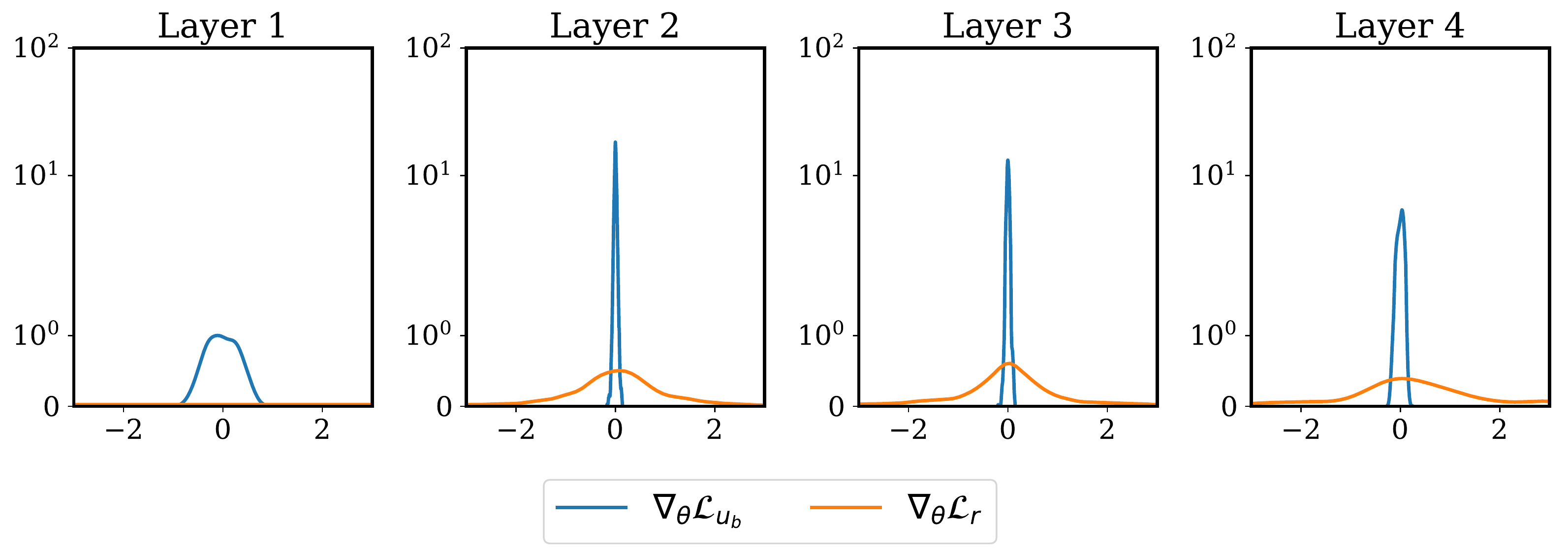}
    \caption{{\em Helmholtz equation:} Histograms of back-propagated gradients $\nabla_\theta \mathcal{L}_r(\theta)$ and $\nabla_\theta \mathcal{L}_{u_b}(\theta)$ at each layer during the $40,000$th iteration of training a standard PINN model for solving the Helmholtz equation.}
    \label{fig:Helmholtz_gradients}
\end{figure}

To explore the reason why this model fails to return accurate predictions, we draw motivation from the seminal work of Glorot and Bengio \cite{glorot2010understanding} and monitor the distribution of the back-propagated gradients of our loss with respect to the neural network parameters during training. Rather than tracking the gradients of the aggregate loss $\mathcal{L}(\theta)$, we track the gradients of each individual terms $\mathcal{L}_{u_b}(\theta) $ and $\mathcal{L}_r(\theta)$ with respect to the weights in each hidden layer of the neural network. As illustrated in figure \ref{fig:Helmholtz_gradients}, the gradients corresponding to the boundary loss term  $\mathcal{L}_{u_b}(\theta)$ in each layer are sharply concentrated around zero and overall attain significantly smaller values than the gradients corresponding to the PDE residual loss $\mathcal{L}_r(\theta)$. As we know, in lack of proper restrictions such as boundary conditions or initial conditions, a PDE system may have infinitely many solutions \cite{evans1998partial}. Therefore, if the gradients $\nabla_\theta \mathcal{L}_{u_b}(\theta)$ are very small during training, then our PINN model should experience difficulties in fitting the boundary conditions. One the other hand, while the gradients $\nabla_\theta \mathcal{L}_r(\theta)$ are large, the neural network can easily learn any solutions that satisfy the equation. As a result, our trained model is heavily biased towards returning a solution that leads to a small PDE residual, but without however respecting the given boundary conditions, it is prone to return erroneous predictions. 

\subsection{Gradient analysis for physics-informed neural networks}


It is now natural to ask what is the mechanism that gives rise to this gradient imbalance between the two loss terms, $\mathcal{L}_{u_b}(\theta) $ and $\mathcal{L}_r(\theta)$, respectively. For starters, one may be quick to notice that the 
chain rule for computing the gradients of $\mathcal{L}_r(\theta)$ is deeper than the chain rule for computing the gradients of $\mathcal{L}_b(\theta)$, and thus the gradients of $\mathcal{L}_r$ are easier to suffer from vanishing gradient pathologies. Paradoxically, here we observe the opposite behavior as it is the $\nabla_\theta \mathcal{L}_{u_b}(\theta)$ gradients that appears to vanish during model training (see figure \ref{fig:Helmholtz_gradients}). To understand what gives rise to this behavior, let us take a step back and consider an even simpler benchmark, the one-dimensional Poisson equation
\begin{align}
\label{eq:Poisson}
    \begin{split}
    &  \Delta u(x) = g(x), \ \ x \in [0,1] \\
    & u(x) = h(x), \ \ x=0 \textrm{ and } x=1.
    \end{split}
\end{align}
To this end, let us consider exact solutions of the form $u(x) = \sin(Cx)$, corresponding to a source term given by $g(x) = - C^2 \sin(Cx) = -C^2 u(x)$. Under the PINNs framework, we will use a fully-connected deep neural network $f_{\theta}(x)$ parametrized by $\theta$ to approximate the latent solution $u(x)$. Then the corresponding loss function over a collection of boundary and residual data-points is given by
\begin{align}
    \begin{split}
         \mathcal{L}(\theta) &=   \mathcal{L}_r(\theta) +  \mathcal{L}_{u_b}(\theta) \\
             &= \frac{1}{N_b} \sum_{i=1}^{N_b}[f_{\theta}(x_b^i) - h(x_b^i)]^2 + \frac{1}{N_r} \sum_{i=1}^{N_r}[\frac{\partial^2}{\partial x^2} f_{\theta}(x_r^i) - g(x_r^i)]^2.
    \end{split}
\end{align}

\begin{figure}
\begin{subfigure}{1.0\textwidth}
  \centering
  \includegraphics[width=.8\linewidth]{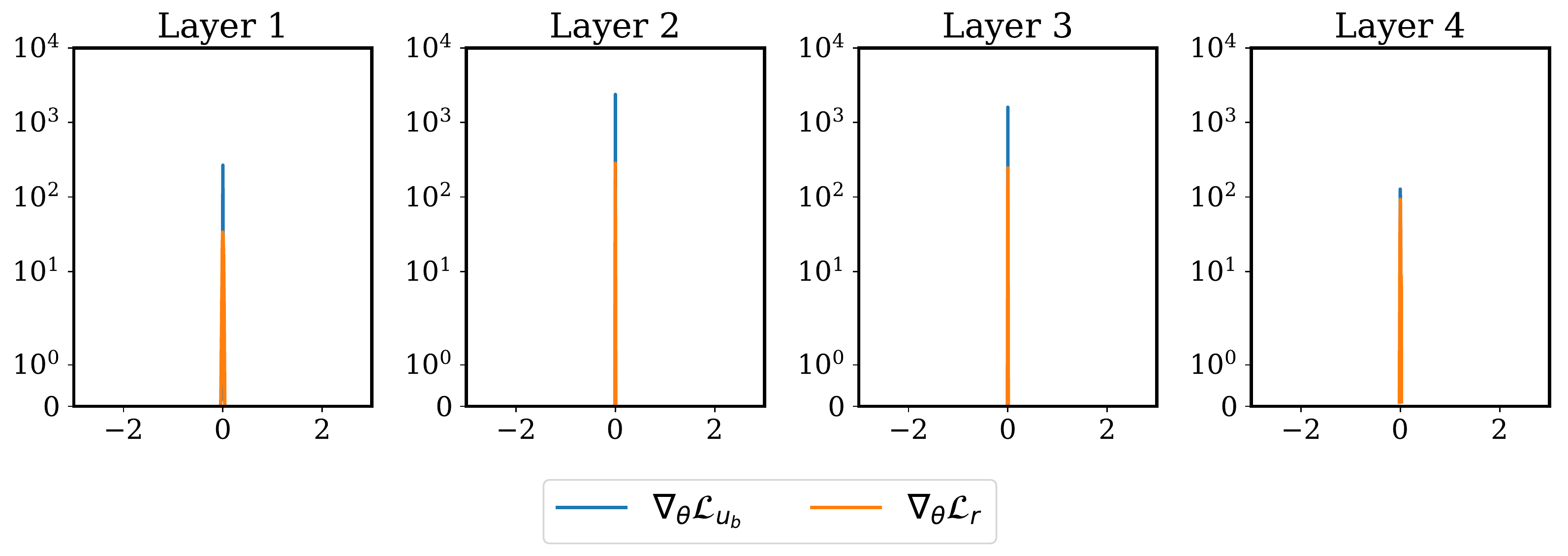}
  \caption{C = 1}
  \label{fig:C1}
\end{subfigure}
\begin{subfigure}{1.0\textwidth}
  \centering
  \includegraphics[width=.8\linewidth]{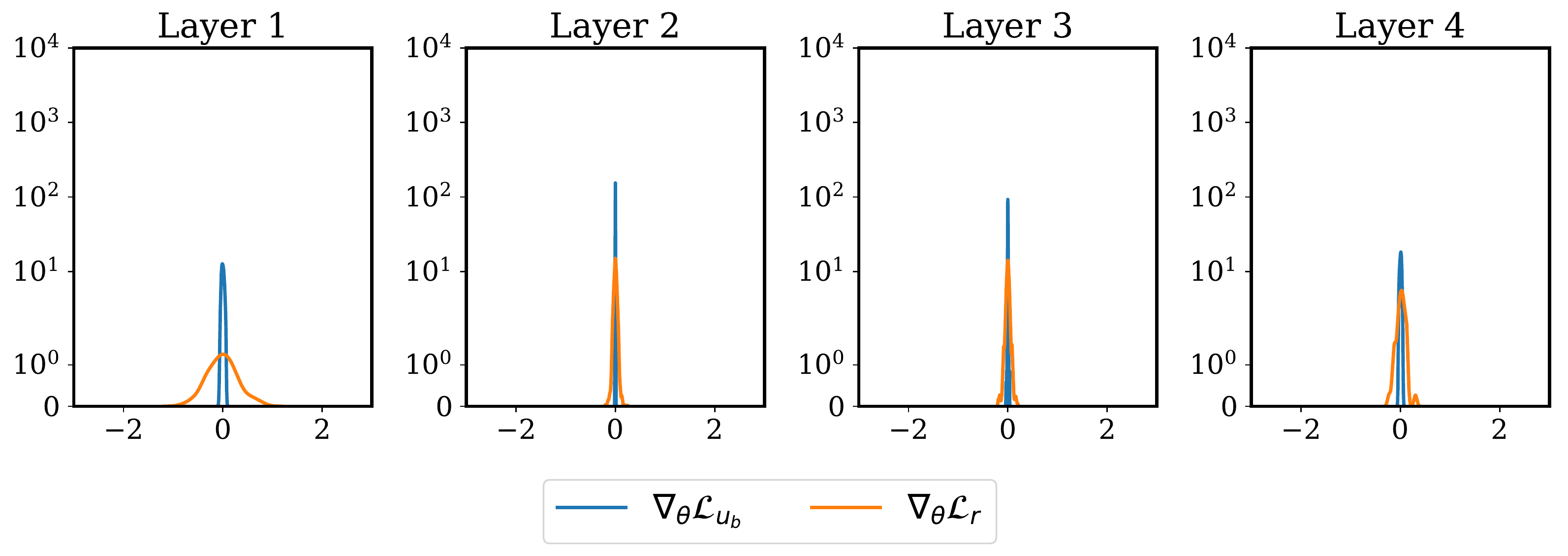}
  \caption{C = 2}
  \label{fig:C2}
\end{subfigure}
\begin{subfigure}{1.0\textwidth}
  \centering
  \includegraphics[width=.8\linewidth]{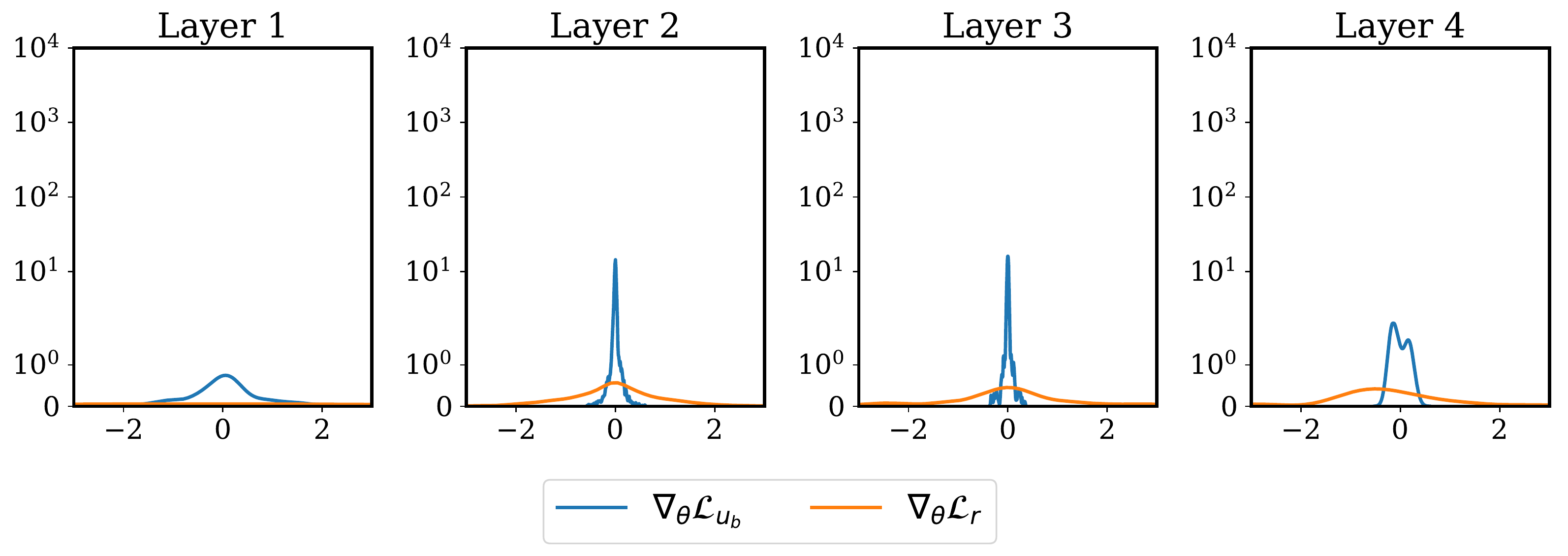}
  \caption{C = 4}
  \label{fig:C4}
\end{subfigure}
\caption{{\em Possion equation:} Histograms of back-propagated gradients $\nabla_\theta \mathcal{L}_r(\theta)$ and $\nabla_\theta \mathcal{L}_{u_b}(\theta)$ at each layer during the $40,000$th iteration of training a standard PINN model for solving the one-dimensional Poisson  equation for different values of the constant $C$, see equation \ref{eq:Poisson}.}
\label{fig:Poisson_benchmark_C}
\end{figure}

To provide some analysis, let us assume that there exists a trained neural network $f_{\theta}(x)$ that can provide a good approximation to the latent solution $u(x)$ (i.e. the target solution is in the Hilbert space spanned by the neural network degrees of freedom). Then we may express our approximation as $f_{\theta}(x) = u(x)\epsilon_{\theta}(x)$,  where $\epsilon_\theta(x)$ is a smooth function defined in $[0,1]$, and for which $|\epsilon_\theta(x) -1| \leq  \epsilon$ for some $\epsilon > 0$, and $\|\frac{\partial^k \epsilon_\theta(x)}{\partial x^k}\|_{L^\infty} < \epsilon$, for all non-negative integer $k$. This construction allows us to derive the following bound for the gradients of the PDE boundary loss and residual loss (see proof in Appendix \ref{sec:appendix_proof})

\begin{align}
\label{eq:gradients_bound}
     &\|\nabla_\theta \mathcal{L}_{u_b}(\theta)\|_{L^\infty} \leq  2 \epsilon \cdot \|\nabla_\theta \epsilon_{\theta}(x)\|_{L^\infty} \\
    &\|\nabla_\theta \mathcal{L}_r(\theta)\|_{L^\infty} \leq O(C^4)  \cdot \epsilon \cdot \|\nabla_\theta \epsilon_{\theta}(x)\|_{L^\infty}
\end{align}
Based on this simple analysis, we can conclude that if the constant $C$ is large, then the norm of gradients of $\mathcal{L}_r(\theta)$ may be much greater the gradients of  $\mathcal{L}_{u_b}(\theta)$, thus biasing the neural network training towards neglecting the contribution of the boundary data-fit term. To confirm this result we have performed a series of simulations in which standard PINN models are trained to approximate the solution of equation \ref{eq:Poisson} for different values of the constant $C$. Our results are summarized in figure \ref{fig:Poisson_benchmark_C}, indicating that larger values of the constant $C$ lead to a pronounced imbalance in the back-propagated gradients $\nabla_\theta \mathcal{L}_r(\theta)$ and $\nabla_\theta \mathcal{L}_{u_b}(\theta)$, that ultimately leads to an inaccurate reconstruction of the PDE solution.

\subsection{Stiffness in the gradient flow dynamics}
\label{sec:stiffness}

To provide further insight into the pathology of imbalanced gradients, let us consider the continuous limit of the learning dynamics for calibrating the neural network parameters $\theta$, as governed by the gradient flow system
\begin{equation}
\label{eq:gradient_flow}
\frac{d\theta}{dt} = -\nabla_\theta \mathcal{L}_r(\theta) - \sum_{i=1}^M \nabla_\theta \mathcal{L}_i(\theta).
\end{equation}
Starting from an initial guess, one can integrate over the gradient flow to obtain a local minimum of the total loss $\mathcal{L}(\theta)$. It is straightforward to see that the gradient descent updates of equation \ref{eq:gradient_descent} correspond to a forward Euler discretization  \cite{iserles2009first} of equation \ref{eq:gradient_flow}. It is also well understood that the stability of this explicit, first-order discretization strategy is severely limited by the choice of the step-size/learning rate, especially for cases in which the governing gradient flow dynamics are stiff. We believe that such cases arise routinely in the context of physics-informed neural networks in which the different terms in their loss have inherently different nature and often correspond to competing objectives (e.g., fitting a set of noisy data versus enforcing a zero PDE residual). 

To obtain a quantitative assessment of stiffness in the gradient flow dynamics of a neural network one can compute and monitor the largest eigenvalue $\sigma_{\max}(\nabla_{\theta}^2\mathcal{L}(\theta))$ of the Hessian matrix $\nabla_{\theta}^2\mathcal{L}(\theta)$ during model training \cite{sagun2016eigenvalues}. This immediately follows from performing a stability analysis for the linearized system
\begin{align}
    \frac{d}{dt}\Tilde{\theta}(t) = -\nabla_{\theta}^2 \mathcal{L}(\Tilde{\theta}(t)) \cdot \Tilde{\theta}(t).
\end{align}
It is well known that the largest eigenvalue of the Hessian dictates the fastest time-scale of the system and directly imposes a restriction on the learning rate that one needs to employ to ensure stability in the forward Euler discretization, as reflected by gradient descent update rule of equation \ref{eq:gradient_descent}. In fact, a classical result in numerical analysis on the conditional stability of the forward Euler method requires bounding the learning rate as $\eta<2/\sigma_{\max}(\nabla_{\theta}^2\mathcal{L}(\theta))$ \cite{iserles2009first}.

\begin{figure}
    \centering
    \includegraphics[width = 0.5\textwidth]{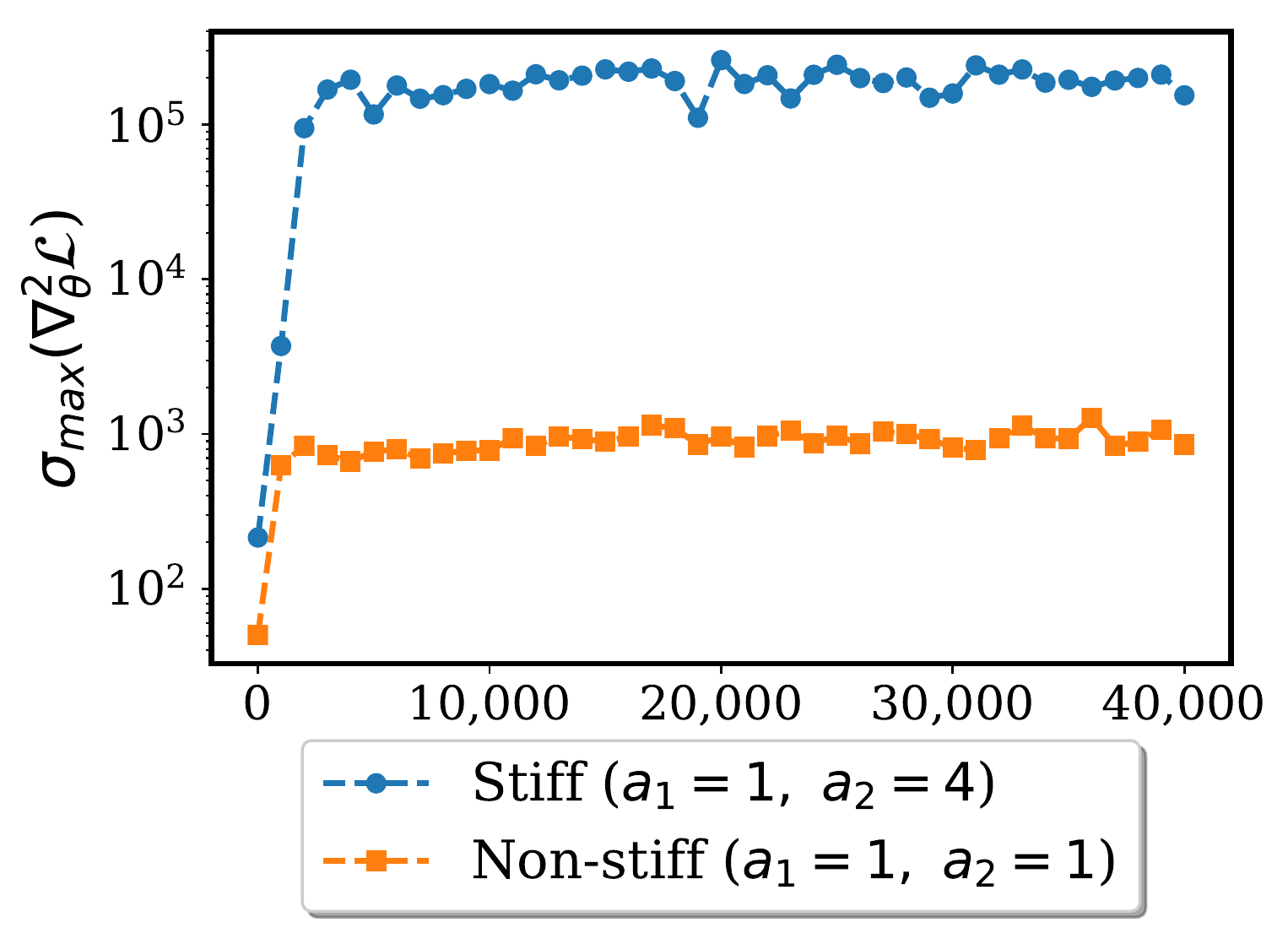}
    \caption{{\em Stiffness in the gradient flow dynamics:} Largest eigenvalue of the Hessian  $\nabla_{\theta}^2\mathcal{L}(\theta)$ during the training of a physics-informed neural network model for approximating the solution to the two-dimensional Helmholtz problem (see equation \ref{eq:Helmholtz}) and for different values of the control parameters $a_1$ and $a_2$.}
    \label{fig:Hessian_eig}
\end{figure}

To illustrate the presence of stiffness in the gradient flow dynamics of physics-informed neural networks, let us revisit the simple simulation study presented in section \ref{sec:gradient_pathologies} for the two-dimensional Helmholtz benchmark. Specifically, we consider two cases by modulating the control parameters $a_1$ and $a_2$ in equation \ref{eq:Helmholtz}. In the first case we set $a_1=1$, $a_2=1$ to fabricate a relatively simple solution $u(x,y)$ that has an isotropic length-scale across both spatial dimensions. In the second case we choose $a_1=1$, $a_2=4$ to fabricate a solution $u(x,y)$ that exhibits directional anisotropy and has more oscillations along the $y$-coordinate. For each case we then employ the same simulation setup presented in section \ref{sec:gradient_pathologies} to train a 4-layer deep physics-informed neural network for approximating the latent solution $u(x,y)$. During training we probe the stiffness of the gradient flow dynamics by computing and monitoring the largest eigenvalue of the Hessian matrix $\sigma_{\max}(\nabla_{\theta}^2\mathcal{L}(\theta))$. In figure \ref{fig:Hessian_eig} we present the recorded values of $\sigma_{\max}(\nabla_{\theta}^2\mathcal{L}(\theta))$ at different iterations of the gradient descent algorithm. Our results reveal that, as the complexity of the target solution is increased, the training dynamics of the physics-informed neural network become increasingly stiffer, ultimately leading to a severe restriction in the required learning rate needed for stable gradient descent updates.

The following simple analysis may reveal the connection between stiffness in the gradient flow dynamics and the evident difficulty in training physics-informed neural networks with gradient descent. Suppose that at $n$-th step of gradient decent during the training,  we have
\begin{equation}
    \label{eq: step n}
    \theta_{n+1} = \theta_n - \eta \nabla_\theta \mathcal{L}(\theta_n) =  \theta_n - \eta      \left[\nabla_\theta \mathcal{L}_r(\theta_n) + \nabla_\theta \mathcal{L}_{u_b}(\theta_n) \right]
\end{equation}
where $\eta$ is the learning rate. 
Then applying second order Taylor expansion to the loss function $\mathcal{L}(\theta)$ at $\theta_n$ gives
\begin{align}
    \label{eq: taylor expansion}
    \mathcal{L}(\theta_{n+1}) =    \mathcal{L}(\theta_{n}) +  (\theta_{n+1} - \theta_{n}) \cdot \nabla_\theta \mathcal{L}(\theta_n)   + \frac{1}{2} (\theta_{n+1} - \theta_{n})^T \nabla^2_\theta \mathcal{L}(\xi)(\theta_{n+1} - \theta_{n})
\end{align}
where $\xi = t\theta_n +  (1-t)\theta_{n+1}$ for some $t \in [0,1]$ and $\nabla^2_\theta \mathcal{L}(\xi)$ is the Hessian matrix of the loss function $\mathcal{L}(\theta)$ evaluated at $\xi$.
Now applying \ref{eq: step n} to \ref{eq: taylor expansion}, we obtain
\begin{align}
    \mathcal{L}(\theta_{n+1}) - \mathcal{L}(\theta_{n}) &= - \eta \nabla_\theta \mathcal{L}(\theta_n) \cdot \nabla_\theta \mathcal{L}(\theta_n) + \frac{1}{2}  \eta^2 \nabla_\theta \mathcal{L}(\theta_n)^T \nabla^2_\theta \mathcal{L}(\xi) \eta \nabla_\theta \mathcal{L}(\theta_n) \\
    &= - \eta \| \nabla_\theta \mathcal{L}(\theta_n) \|_2^2 + \frac{1}{2}  \eta^2 \nabla_\theta \mathcal{L}(\theta_n)^T \nabla^2_\theta \mathcal{L}(\xi) \nabla_\theta \mathcal{L}(\theta_n) \\
    &= - \eta \| \nabla_\theta \mathcal{L}(\theta_n) \|_2^2 + \frac{1}{2}  \eta^2 \nabla_\theta \mathcal{L}(\theta_n)^T \left(\nabla^2_\theta \mathcal{L}_r(\xi) + \nabla^2_\theta \mathcal{L}_{u_b}(\xi)  \right)\nabla_\theta \mathcal{L}(\theta_n) \\
    &=  - \eta \| \nabla_\theta \mathcal{L}(\theta_n) \|_2^2 + \frac{1}{2}  \eta^2 \nabla_\theta \mathcal{L}(\theta_n)^T \nabla^2_\theta \mathcal{L}_r(\xi) \nabla_\theta \mathcal{L}(\theta_n) + \frac{1}{2}  \eta^2 \nabla_\theta \mathcal{L}(\theta_n)^T \nabla^2_\theta \mathcal{L}_{u_b}(\xi) \nabla_\theta \mathcal{L}(\theta_n)
\end{align} 
Here, note that
\begin{align}
\nabla_\theta \mathcal{L}(\theta_n)^T \nabla^2_\theta \mathcal{L}(\xi)  \nabla_\theta \mathcal{L}(\theta_n) &= \| \nabla_\theta \mathcal{L}(\theta_n)  \|_2^2 \frac{\nabla_\theta \mathcal{L}(\theta_n)^T}{\| \nabla_\theta \mathcal{L}(\theta_n)  \|}  \nabla^2_\theta \mathcal{L}(\xi)  \frac{\nabla_\theta \mathcal{L}(\theta_n)}{\| \nabla_\theta \mathcal{L}(\theta_n)  \|} \\
& = \| \nabla_\theta \mathcal{L}(\theta_n)  \|_2^2 x^T Q^T \text{diag}(\lambda_1, \lambda_2 \cdots \lambda_n) Q x \\
&=  \| \nabla_\theta \mathcal{L}(\theta_n)  \|_2^2 y^T \text{diag}(\lambda_1, \lambda_2 \dots \lambda_M) y \\
&= \| \nabla_\theta \mathcal{L}(\theta_n)  \|_2^2 \sum_{i=1}^M \lambda_i y_i^2 
\end{align}
where $x =  \frac{\nabla_\theta \mathcal{L}(\theta_n)}{\| \nabla_\theta \mathcal{L}(\theta_n)  \|}$, $Q$ is an orthogonal matrix diagonalizing $\nabla^2_\theta \mathcal{L}(\xi)$ and $y = Qx$. And $\lambda_1 \leq  \lambda_2 \leq \cdots \leq  \lambda_N$ are eigenvalues of $\nabla^2_\theta \mathcal{L}(\xi)$. 
Similarly, we have
\begin{align}
    &\nabla_\theta \mathcal{L}(\theta_n)^T \nabla^2_\theta \mathcal{L}_r(\xi)  \nabla_\theta \mathcal{L}(\theta_n) = \| \nabla_\theta \mathcal{L}(\theta_n)  \|_2^2 \sum_{i=1}^M \lambda_i^r y_i^2  \\
    &    \nabla_\theta \mathcal{L}(\theta_n)^T \nabla^2_\theta \mathcal{L}_{u_b}(\xi)  \nabla_\theta \mathcal{L}(\theta_n) = \| \nabla_\theta \mathcal{L}(\theta_n)  \|_2^2 \sum_{i=1}^M \lambda_i^{u_b} y_i^2 
\end{align}
where $\lambda_1^r \leq  \lambda_2^r \leq \cdots \leq  \lambda_N^r$ and  $\lambda_1^{u_b} \leq  \lambda_2^{u_b} \leq \cdots \leq  \lambda_N^{u_b}$
are eigenvalues of $\nabla^2_\theta \mathcal{L}_r$ and $\nabla^2_\theta \mathcal{L}_{u_b}$ respectively.
Thus, combining these together we get 
\begin{align}
    &\mathcal{L}(\theta_{n+1}) - \mathcal{L}(\theta_{n}) = \eta  \| \nabla_\theta \mathcal{L}(\theta_n)  \|_2^2 (-1 + \frac{1}{2 } \eta \sum_{i=1}^N \lambda_i y_i^2 ) \\
    &\mathcal{L}_r(\theta_{n+1}) - \mathcal{L}_r(\theta_{n}) = \eta  \| \nabla_\theta \mathcal{L}(\theta_n)  \|_2^2 (-1 + \frac{1}{2 } \eta \sum_{i=1}^N \lambda_i^r y_i^2 ) \\
      &\mathcal{L}_{u_b}(\theta_{n+1}) - \mathcal{L}_{u_b}(\theta_{n}) = \eta  \| \nabla_\theta \mathcal{L}(\theta_n)  \|_2^2 (-1 + \frac{1}{2 } \eta \sum_{i=1}^N \lambda_i^{u_b} y_i^2 )
\end{align}

Thus, if the gradient flow is stiff, then many eigenvalues of $\nabla^2_\theta \mathcal{L}(\xi)$ may be large. As a result, it is very possible that $ \mathcal{L}(\theta_{n+1}) - \mathcal{L}(\theta_{n}) >0$, which implies that the gradient decent method fails to decrease the loss even if $\nabla_\theta \mathcal{L}( \theta_n)$ is the right decent direction.  This result comes to no surprise, as it is well understood that gradient descent may get stuck in limit cycles or even diverge in the presence of multiple competing objectives \cite{mertikopoulos2018cycles,balduzzi2018mechanics}.

\begin{figure}
     \centering
     \begin{subfigure}[b]{0.4\textwidth}
         \centering
         \includegraphics[width=\textwidth]{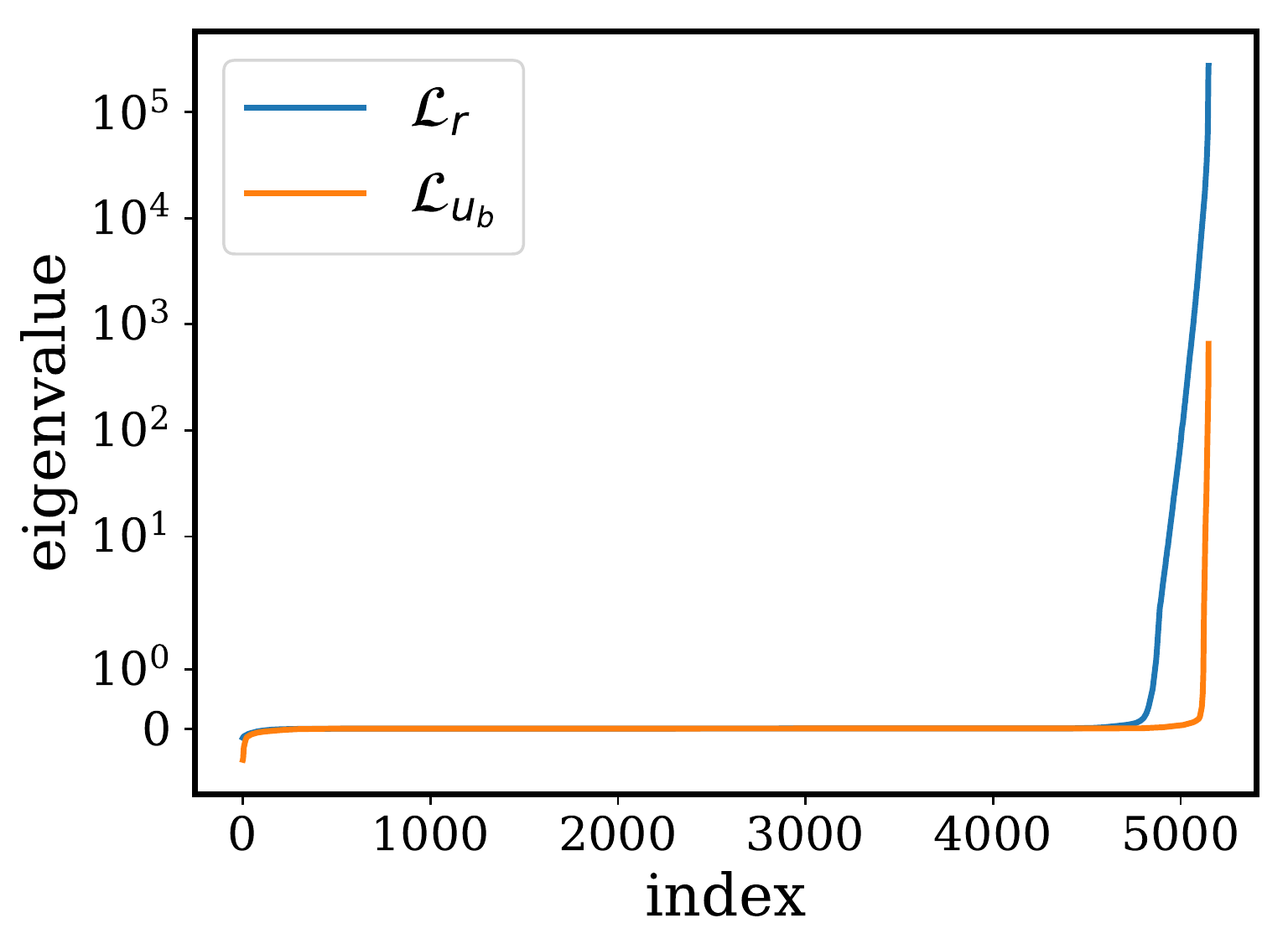}
         \caption{}
         \label{fig: eigenvalues}
     \end{subfigure}
     \begin{subfigure}[b]{0.4\textwidth}
         \centering
         \includegraphics[width=\textwidth]{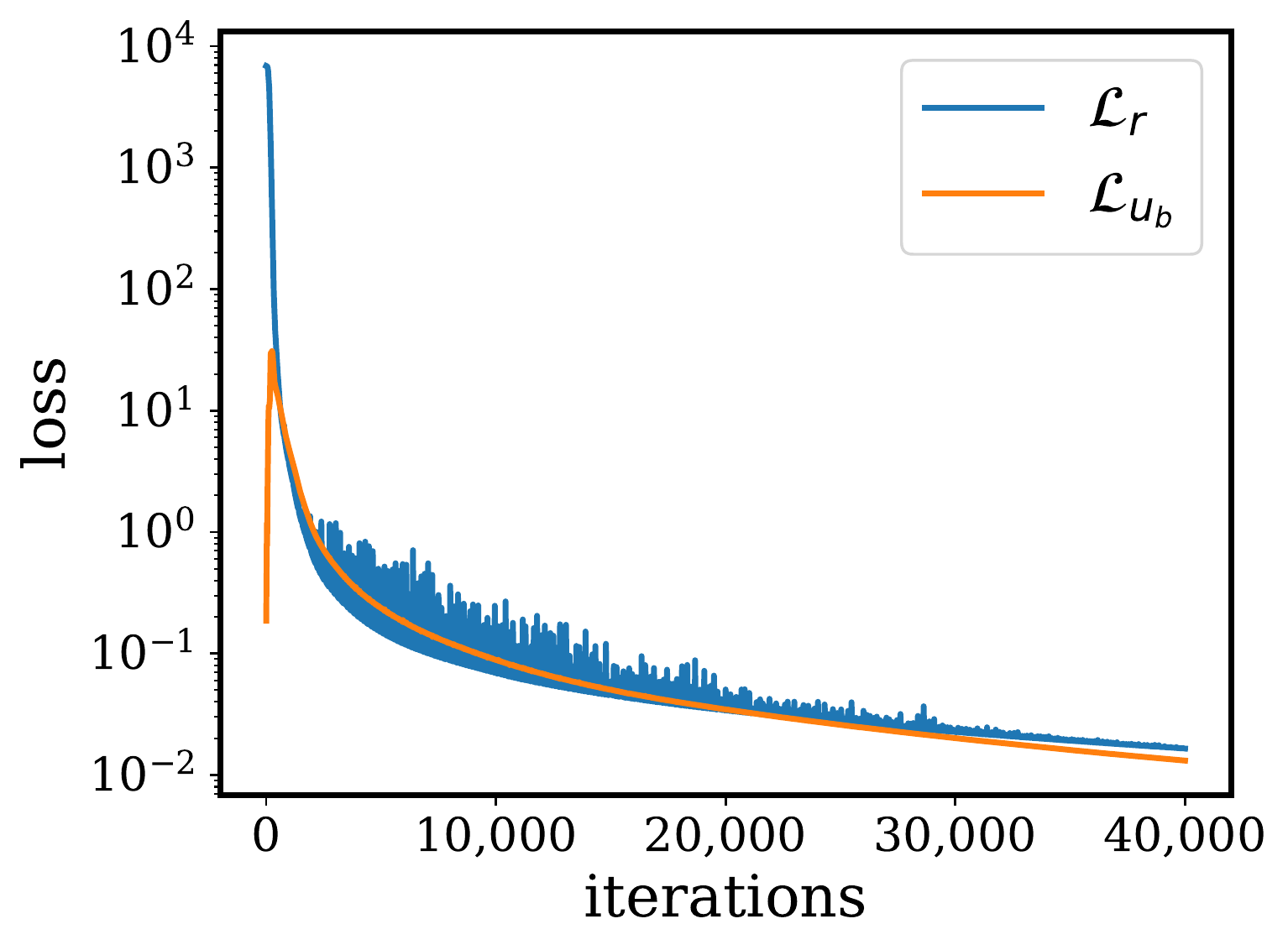}
         \caption{}
         \label{fig: loss}
     \end{subfigure}
    \caption{{\em Helmholtz equation:} (a) All eigenvalues of $\nabla^2_\theta \mathcal{L}_r (\theta)$ and $\nabla^2_\theta \mathcal{L}_{u_b} (\theta)$, respectively, arranged in increasing order. (b) Loss curves of $\mathcal{L}_r(\theta)$ and $\mathcal{L}_{u_b}$, respectively, after 40,000 iterations of gradient decent using the Adam optimizer in full batch mode.} 
\end{figure}

To verify our hypothesis, we still consider the example in section \ref{sec:methods}. Specifically, we choose a five layer neural network with 50 units in each hidden layer. To reduce stochastic effects introduced by mini-batch gradient descent we use full batch gradient decent with an initial learning rate of $10e-3$ and an exponential decay with a decay-rate of 0.9 and a decay-step of 1,000 iterations. All collocation points are uniformly sampled from both the boundary and the interior domain. Figure \ref{fig: eigenvalues} shows the eigenvalues of $\nabla^2_\theta \mathcal{L}_{u_b}(\theta)$ and $\nabla^2_\theta \mathcal{L}_r(\theta)$ in ascending order. Clearly, we can conclude that many eigenvalues of $\nabla^2_\theta \mathcal{L}_r(\theta)$ are extremely large up to $10^5$ and the stiffness of the gradient flow is dominated by $\mathcal{L}_{r}(\theta)$. As a consequence, in figure \ref{fig: loss} we observe that the loss of $\mathcal{L}_{r}(\theta)$ oscillates wildly even when full batch gradient descent is employed, while the loss of $\mathcal{L}_{u_b}(\theta)$ exhibits a decreasing tendency since the eigenvalues of $\nabla^2_\theta \mathcal{L}_{u_b}(\theta)$ remain relatively small. Besides, note that the oscillation of the loss of $\mathcal{L}_r(\theta)$ becomes small as the iteration increases since the learning rate decay. The above analysis may give another angle to illustrate how the stiff nature of gradient flow system may cause severe difficulties in the training of physics-informed neural networks with gradient descent. It also hints to the importance of choosing the learning rate such us a stable discretization of the gradient flow is achieved..

Moreover, applying the classical Sobolev inequality \cite{adams2003sobolev} to $\nabla_\theta \mathcal{L}_r$ and $\nabla_\theta \mathcal{L}_{u_b}$ gives 
\begin{align}
    \label{eq: soboblev_grad_bound_1}
    &\|\nabla_\theta \mathcal{L}_r(\theta)\|_{L^\infty} \leq C  \|\nabla^2_\theta \mathcal{L}_r(\theta)\|_{L^\infty} \\
    \label{eq: soboblev_grad_bound_2}
    &\|\nabla_\theta \mathcal{L}_{u_b}(\theta)\|_{L^\infty} \leq C  \|\nabla^2_\theta \mathcal{L}_{u_b}(\theta)\|_{L^\infty}
\end{align}
where $C$ is a constant that does not depend on $\nabla \mathcal{L}_r(\theta)$ and $\nabla \mathcal{L}_{u_b}(\theta)$. By diagonalizing $\nabla^2 \mathcal{L}_r(\theta)$ and $\nabla^2 \mathcal{L}_{u_b}(\theta)$ with orthonormal matrices, we know that $\|\nabla^2_\theta \mathcal{L}_{r}(\theta)\|_{L^\infty}$ and $\|\nabla^2_\theta \mathcal{L}_{u_b}(\theta)\|_{L^\infty}$ are bounded by the largest eigenvalues of $\nabla^2 \mathcal{L}_r(\theta)$ and $\nabla^2 \mathcal{L}_{u_b}(\theta)$ respectively. Then again from  figure \ref{fig: eigenvalues}, we can conclude that $\|\nabla^2_\theta \mathcal{L}_r(\theta)\|_{L^\infty}$ is much greater than $\|\nabla^2_\theta \mathcal{L}_{u_b}(\theta)\|_{L^\infty}$. Therefore, equations \ref{eq: soboblev_grad_bound_1}, \ref{eq: soboblev_grad_bound_2} give another loose upper bound for $\nabla \mathcal{L}_r(\theta)$ and  $\nabla \mathcal{L}_{u_b}(\theta)$ and reveal the reason why stiffness of the gradient flow may result in unbalanced gradients pathologies.

Based on these findings we may conclude that the {\em de facto} use of gradient descent and its modern variants (e.g., Adam \cite{kingma2014adam}) may be a poor, or even unstable choice for training physics-informed neural networks. From a numerical analysis standpoint, it is then natural to investigate whether more stable discretizations, such as implicit or IMEX operator splitting schemes \cite{iserles2009first,strang1968construction}, can yield more effective optimization algorithms for constrained neural networks. From an optimization standpoint, this setting has motivated the development of proximal gradient algorithms \cite{parikh2014proximal} that have been successfully applied to ill-posed inverse problems involving stiff regularization terms \cite{mardani2018neural}, as well as effective algorithms for finding Nash equilibria in multi-player games \cite{balduzzi2018mechanics, schafer2019competitive}. Nevertheless, these intriguing directions go beyond the scope of the present work and define an areas for future investigation. As discussed in the next section, here our goal is to retain the simplicity of gradient descent while providing an non-intrusive and adaptive heuristic for enhancing its performance. Similar approaches can be found in the numerical analysis literature for integrating stiff and multi-scale dynamical systems, see for example the multi-rate schemes of Gear {\em et. al.} \cite{gear1984multirate} and the flow-averaging integration techniques put forth by Tao {\it et. al.} \cite{tao2010nonintrusive}. 

\subsection{Learning rate annealing for physics-informed neural networks}
\label{sec:adaptive}

Having identified a common mode of failure for physics-informed neural networks related to unbalanced gradients during back-propagation, we can investigate potential remedies for overcoming this pathology. To do so, let us re-examine the general form of a PINNs loss function
\begin{align}
   \mathcal{L}(\theta) :=  \mathcal{L}_r(\theta) + \sum\limits_{i=1}^{M} \mathcal{L}_i(\theta), 
\end{align}
the minimization of which is typically performed according to the following gradient descent update
\begin{align}
    \theta_{n+1} &= \theta_n - \eta \nabla_\theta \mathcal{L}(\theta_n) \nonumber\\
                 &= \theta_n - \eta [\nabla_\theta \mathcal{L}_r(\theta_n) + \sum_{i=1}^M \nabla_\theta \mathcal{L}_i(\theta_n)]
\end{align}
where $\eta$ is a learning rate parameter. To balance the interplay between the different terms in this loss, a straightforward way is to multiply a constant $\lambda_i$ to each $\mathcal{L}_i(\theta)$ term. More specifically, we consider minimizing the following  loss in which the weights $\lambda$ resemble the role of penalty coefficients in constrained optimization \cite{bertsekas2014constrained}
\begin{align}
    \label{eq:weighted_loss}
   \mathcal{L}(\theta) :=  \mathcal{L}_r(\theta) + \sum\limits_{i=1}^{M}\lambda_i \mathcal{L}_i(\theta). 
\end{align}
Consequently, the corresponding gradient descent updates now take the form
\begin{align}
    \theta_{n+1} &= \theta_n - \eta \nabla_\theta \mathcal{L}(\theta_n) \\
                 &= \theta_n - \eta \nabla_\theta \mathcal{L}_r(\theta_n) - \eta \sum_{i=1}^M  \lambda_i \nabla_\theta \mathcal{L}_i(\theta_n)\label{eq:gradient_descent},
\end{align}
where we see how the constants $\lambda_i$ can effectively introduce a re-scaling of the learning rate corresponding to each loss term. Obviously, the next question that needs to be answered is should those weights how $\lambda_i$ be chosen? It is straightforward to see that choosing $\lambda_i$ arbitrarily following a trial and error procedure is extremely tedious and may not produce satisfying results. Moreover, the optimal constants may vary greatly for different problems, which means we cannot find a fixed empirical recipe that is transferable across different PDEs.  Most importantly, the loss function always consists of various parts that serve to provide restrictions on the equation. It is impractical to give different weights to different parts of the loss function manually.

Here we draw motivation from Adam \cite{kingma2014adam} -- one the most widely used adaptive learning rate optimizers in the deep learning literature -- to derive an adaptive rule for choosing the $\lambda_i$ weights online during model training. The basic idea behind Adam is to keep track of the first- and second-order moments of the back-propagated gradients during training, and utilize this information to adaptively scale the learning rate associated with each parameter in the $\theta$ vector. 
In a similar spirit, 
our proposed learning rate annealing procedure, as summarized in algorithm \ref{alg:learning_rate_annealing}, is designed to automatically tune the $\lambda_i$ weights by utilizing the back-propagated gradient statistics during model training, such that the interplay between all terms in equation \ref{eq:weighted_loss} is appropriately balanced.

\begin{algorithm}
\SetAlgoLined
Consider a physics-informed neural network $f_{\theta}(\bm{x})$ with parameters $\theta$ 
and a loss function 
\begin{align*}
\mathcal{L}(\theta) :=  \mathcal{L}_r(\theta) + \sum\limits_{i=1}^{M}\lambda_i \mathcal{L}_i(\theta), 
\end{align*}
where $\mathcal{L}_r(\theta)$ denotes the PDE residual loss, the $\mathcal{L}_i(\theta)$ correspond to data-fit terms (e.g., measurements, initial or boundary conditions, etc.), and $\lambda_i = 1$, $i=1,\dots,M$ are free parameters used to balance the interplay between the different loss terms. Then use $S$ steps of a gradient descent algorithm to update the parameters $\theta$ as:

 \For{$n = 1, \dots, S$}{
  (a) Compute $\hat{\lambda}_i$ by
  \begin{align}
  \label{eq:lambda_hat_update}
     \hat{\lambda}_i = \dfrac{\max_{\theta}\{|\nabla_{\theta}\mathcal{L}_r(\theta_n)| \}} {\overline{|\nabla_{\theta}\mathcal{L}_i(\theta_n)|}}, \ \ i=1,\dots,M,
  \end{align}
   where $\overline{|\nabla_{\theta}\mathcal{L}_i(\theta_n)|}$ denotes the mean of $|\nabla_{\theta}\mathcal{L}_i(\theta_n)|$ with respect to parameters $\theta$.
   
  (b) Update the weights $\lambda_i$ using a moving average of the form
  \begin{align}
  \label{eq:lambda_update}
      \lambda_i = (1 - \alpha ) \lambda_i + \alpha \hat{\lambda}_i, \ \ i=1,\dots,M.
  \end{align}
  
  (c) Update the parameters $\theta$ via gradient descent
  \begin{align}
  \label{eq:theta_update}
      \theta_{n+1} = \theta_{n} - \eta \nabla_{\theta}\mathcal{L}_r(\theta_n) - \eta \sum\limits_{i=1}^{M}\lambda_i \nabla_{\theta}\mathcal{L}_i(\theta_n)
  \end{align}
}

The recommended hyper-parameter values are: $\eta = 10^{-3}$ and $\alpha=0.9$.

\caption{Learning rate annealing for physics-informed neural networks}
\label{alg:learning_rate_annealing}
\end{algorithm}

The proposed algorithm is characterized by two steps. First, we compute instantaneous values for the constants $\hat{\lambda}_i$ by computing the ratio between the maximum gradient value attained by $\nabla_\theta \mathcal{L}_r(\theta)$ and the mean of the gradient magnitudes computed for each of the $\mathcal{L}_i(\theta)$ loss terms, namely, $\overline{|\nabla_{\theta}\mathcal{L}_i(\theta)|}$, see equation \ref{eq:lambda_hat_update}. As these instantaneous values are expected to exhibit high variance due to the stochastic nature of the gradient descent updates, the actual weights $\lambda_i$ are computed as a running average of their previous values, as shown in equation \ref{eq:lambda_update}. Notice that the updates in equations \ref{eq:lambda_hat_update} and \ref{eq:lambda_update} can either take place at every iteration of the gradient descent loop, or at a frequency specified by the user (e.g., every 10 gradient descent steps). Finally, a gradient descent update takes is performed to update the neural network parameters $\theta$ using the current weight values stored in $\lambda_i$, see equation \ref{eq:theta_update}.

The key benefits of the proposed automated procedure is that this adaptive method can be easily generalized to loss functions consisting of multiple terms (e.g., multi-variate problems with multiple boundary conditions on different variables), while the extra computational overhead associated with computing the gradient statistics in equation \ref{eq:lambda_hat_update} is small, especially in the case of infrequent updates. Moreover, our computational studies confirm very low sensitivity on the additional hyper-parameter $\alpha$, as the accuracy of our results does not exhibit any significant variance when this parameter take values within a reasonable range (e.g., $\alpha\in[0.5,0.9]$).

\begin{figure}
    \centering
    \includegraphics[width = \textwidth]{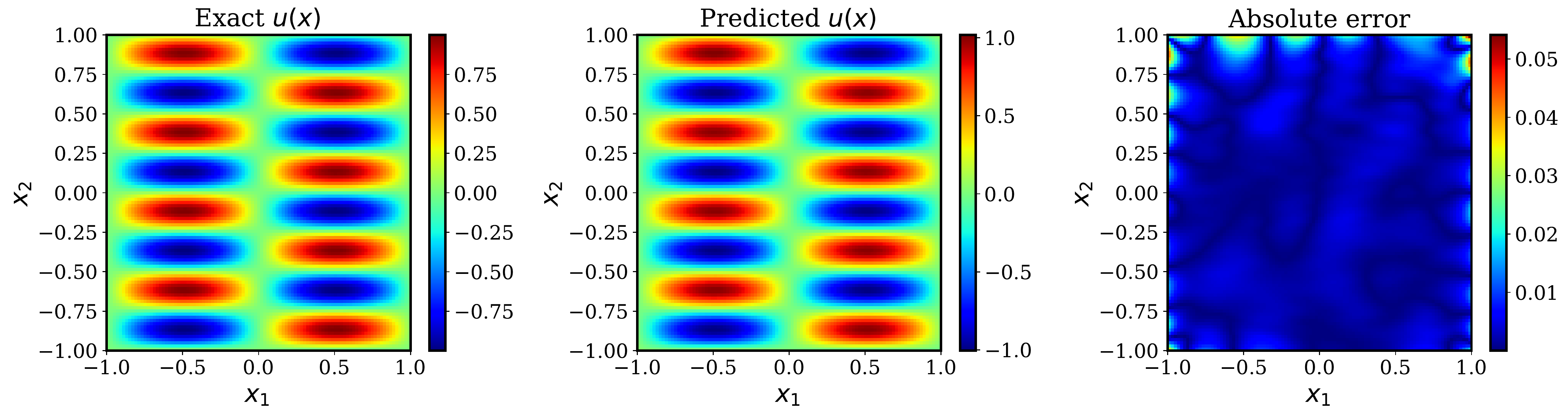}
    \caption{{\em Helmholtz equation:}  Exact solution versus the prediction of a physics-informed neural network model with 4 hidden layers and 50 neurons each layer after 40,000 iterations of training using the proposed learning rate annealing algorithm  (relative $L^2$-error: 1.27e-02).}
    \label{fig:Helmholtz_prediction_adaptive}
\end{figure}

\begin{figure}
    \centering
    \includegraphics[width = \textwidth]{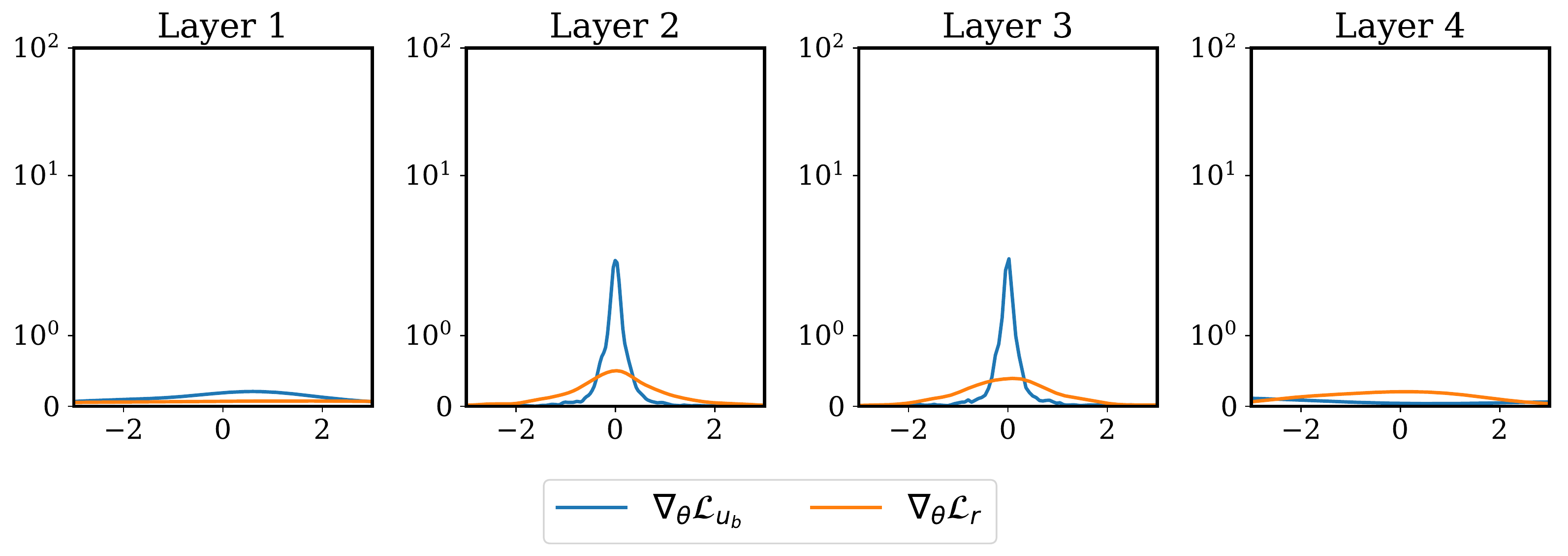}
    \caption{{\em Helmholtz equation:} Histograms of back-propagated gradients of $\nabla_\theta \mathcal{L}_r$ and $\nabla_\theta \mathcal{L}_{u_b}$ at each layer during training a PINN with the proposed algorithm \ref{alg:learning_rate_annealing} to solve Helmholtz equation.}
    \label{fig:Helmholtz_gradients_adaptive}
\end{figure}

\begin{figure}
    \centering
    \includegraphics[width = 0.5\textwidth]{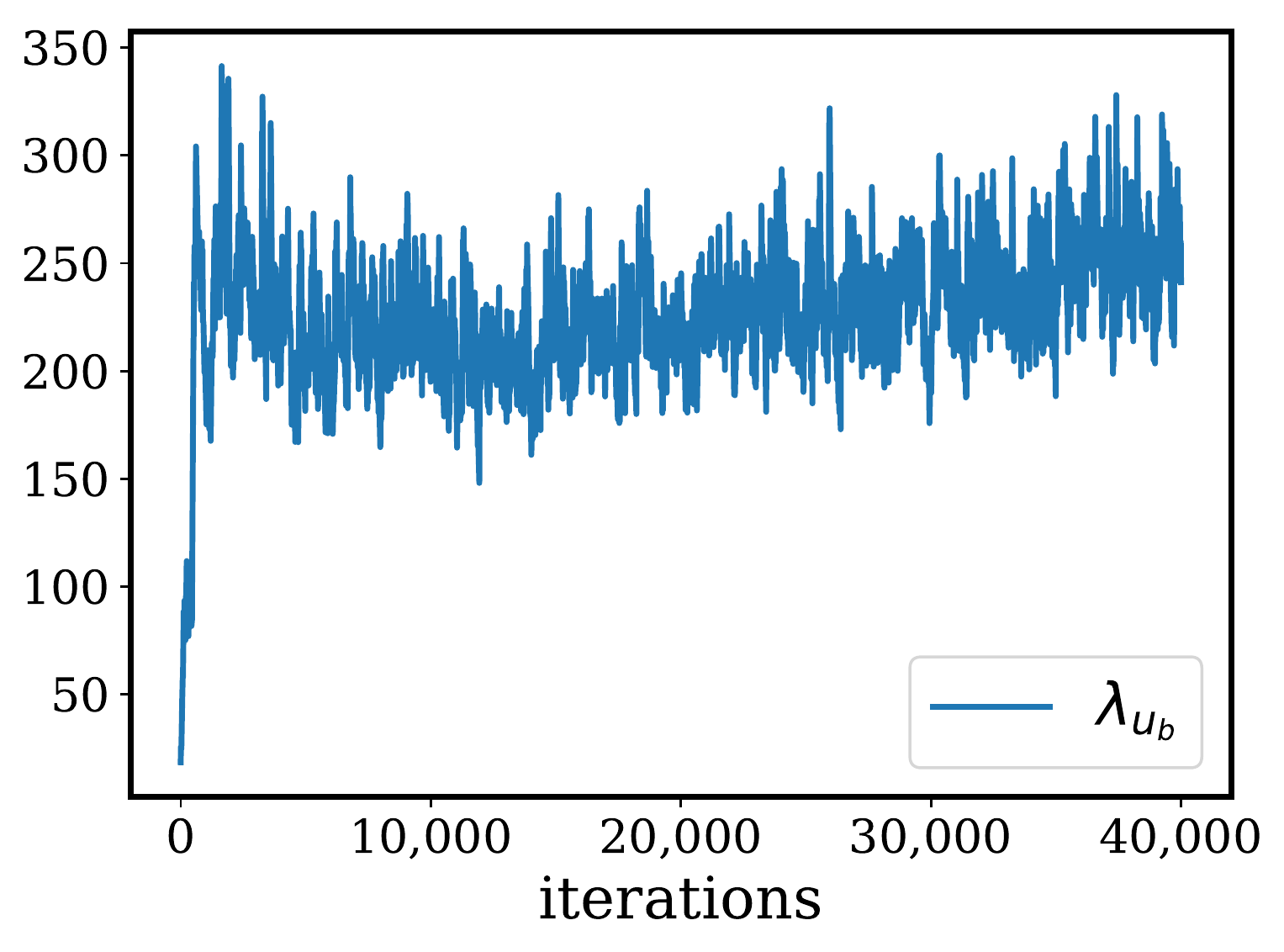}
    \caption{{\em Helmholtz equation:} Evolution of the constant $\lambda_{u_{b}}$ used to scale the boundary condition loss $\mathcal{L}_{u_{b}}(\theta)$ in equation \ref{eq:loss_Helmholtz} during model training using algorithm \ref{alg:learning_rate_annealing}.}
    \label{fig:Helmholtz_prediction_adaptive_constant}
\end{figure}

A first illustration of the effectiveness of the proposed learning-rate annealing algorithm is provided through the lens of the Helmholtz benchmark presented in equation \ref{eq:Helmholtz}. Figures \ref{fig:Helmholtz_prediction_adaptive} and \ref{fig:Helmholtz_gradients_adaptive} summarize the predictions of the same 4-layer deep physics-informed neural network model used in section \ref{sec:gradient_pathologies} after training it using algorithm \ref{alg:learning_rate_annealing} for $40,000$ gradient descent iterations. Evidently, the proposed training scheme is able to properly balance the interplay between the boundary and the residual loss, and improve the relative prediction error by more than one order of magnitude. In particular, by comparing figures \ref{fig:Helmholtz_prediction} and \ref{fig:Helmholtz_prediction_adaptive} notice how the absolute point-wise error at the domain boundaries has been significantly reduced. Finally, figure \ref{fig:Helmholtz_prediction_adaptive_constant} summarizes the convergent evolution of the constant $\lambda_{u_{b}}$ used to scale the boundary condition loss $\mathcal{L}_{u_{b}}(\theta)$ in equation \ref{eq:loss_Helmholtz} during model training using algorithm \ref{alg:learning_rate_annealing}.

\subsection{An improved fully-connected neural architecture}
\label{sec:improved_FC}

Besides carefully formulating a loss function (and effective algorithms to minimize it), another ingredient that is key to the success of deep neural networks is the design of their architecture. The most successful neural architectures in the literature are in principle designed to exploit prior knowledge associated with a given task. For instance, convolutional neural networks \cite{krizhevsky2012imagenet,goodfellow2016deep} are widely used for object recognition and image classification tasks. This is because the convolution operation naturally adheres to the symmetries induced by the translation group, which is a crucial feature in image recognition. Similarly, recurrent neural networks \cite{lipton2015critical,goodfellow2016deep} are well suited to modeling sequence data due to their ability to respect temporal invariance and capture long-term dependencies. Although respecting the invariances that characterize a given task is crucial in designing an effective neural architecture, in many scenarios related to physical systems the underlying symmetry groups are often non-trivial or even unknown. This is the main reason why the majority of recent works focused on applying deep learning tools to modeling and simulating physical systems governed by PDEs, still relies on generic fully connected architectures \cite{raissi2019physics,tripathy2018deep,sirignano2018dgm,tartakovsky1808learning,sun2019surrogate}.
Examples of cases include recent studies on physics-informed neural networks \cite{raissi2019physics, raissi2018multistep}, which primarily use fully-connected architectures with weak inductive biases and simply rely on the fact that the representational capacity of these over-parametrized networks can be sufficient to correctly capture the target solution of a given PDE. However, in absence of a universal approximation theorem for physics-informed neural networks (i.e., fully-connected networks trained subject to PDE constraints), it is currently unclear whether standard fully-connected architectures offer sufficiently flexible representations for inferring the solution of complex PDEs. Although this question remains hard to answer, in this section we show how a simple extension to standard fully-connected networks can lead to a novel architecture that appears to perform uniformly and significantly better across all the benchmark studies considered in this work.

\begin{figure}
    \centering
    \includegraphics[width = 0.7\textwidth]{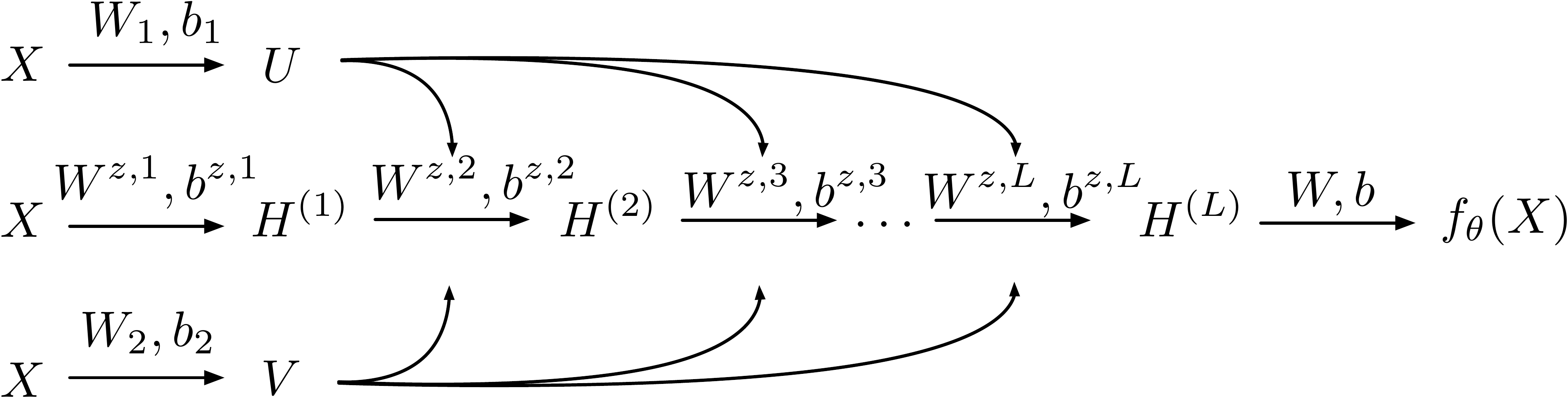}
    \caption{{\em An improved fully-connected architecture for physics-informed neural networks:} Introducing residual connections and accounting for multiplicative interactions between the inputs can lead to improved predictive performance.}
    \label{fig:ADGM}
\end{figure}

Inspired by neural attention mechanisms recently employed for computer vision and natural language processing tasks \cite{vaswani2017attention}, here we present a novel neural network architecture, which has the following characteristics: (i) explicitly accounts for multiplicative interactions between different input dimensions, and (ii) enhances the hidden states with residual connections. As shown in figure \ref{fig:ADGM} the key extension to conventional fully-connected architectures is the introduction of two transformer networks that project the inputs variables to a high-dimensional feature space, and then use a point-wise multiplication operation to update the hidden layers according to the following forward propagation rule

\begin{align}
    &U = \phi(X W^1 + b^1), \ \  V = \phi(X W^2 + b^2) \\
    &H^{(1)} = \phi(X W^{z,1} + b^{z, 1}) \\
    &Z^{(k)} = \phi(H^{(k)}W^{z,k} + b^{z, k}), \ \ k=1, \dots, L \\
    &H^{(k+1)} = (1 - Z^{(k)}) \odot U  +  Z^{(k)}  \odot V, \ \  k=1, \dots, L \\
   & f_{\theta}(x) = H^{(L+1)}W  + b
\end{align}
where $X$ denotes the $(n\times d)$ design matrix of the input data-points, and $\odot$ denotes element-wise multiplication. The parameters of this model are essentially the same as in a standard fully-connected architecture, with the addition of the weights and biases used by the two transformer networks, i.e.,
\begin{align}
    \theta = \{W^1, b^1, W^2, b^2, (W^{z,l}, b^{z,l})_{l=1}^L, W,b  \}
\end{align}
Finally, the number of units in each layer is $M$ and $\phi : \mathbb{R}^M \rightarrow \mathbb{R}^M$ is a nonlinear activation function. Here we should also notice that the proposed architecture, and its associated forward pass, induce a relatively small computational and memory overhead while leading to significant improvements in predictive accuracy. 

\begin{figure}
    \centering
    \includegraphics[width = \textwidth]{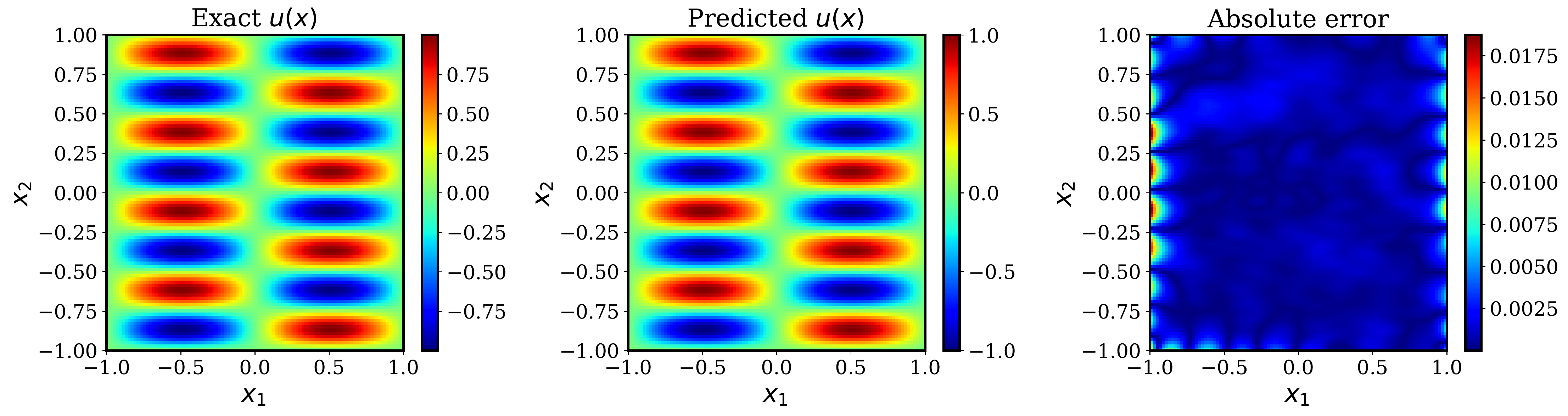}
    \caption{{\em Helmholtz equation:} Exact solution versus the prediction of a physics-informed neural network model with the proposed improved architecture (4 hidden layers, 50 neurons each) after 40,000 iterations of training using the proposed learning rate annealing algorithm  (relative $L^2$-error: 3.69e-03).}
    \label{fig:Helmholtz_prediction_adaptive_ADGM}
\end{figure}

A first indication of the improved performance brought by the proposed neural architecture can be seen by revisiting the Helmholtz benchmark discussed in section \ref{sec:gradient_pathologies}. Figure \ref{fig:Helmholtz_prediction_adaptive_ADGM} summarizes the prediction of the proposed fully connected architecture with a depth of 4 layers for a physics-informed neural network model trained it using algorithm \ref{alg:learning_rate_annealing} for $40,000$ gradient descent iterations. Evidently, the proposed training scheme is able to properly balance the interplay between the boundary and the residual loss, and improve the relative prediction error by almost two orders of magnitude compared to the original formulation of Raissi {\em et. al.} \cite{raissi2019physics}. In particular, by comparing figures \ref{fig:Helmholtz_prediction} and \ref{fig:Helmholtz_prediction_adaptive_ADGM} notice how the absolute point-wise error at the domain boundaries has been effectively diminished.

\begin{figure}
    \centering
    \includegraphics[width = 0.5\textwidth]{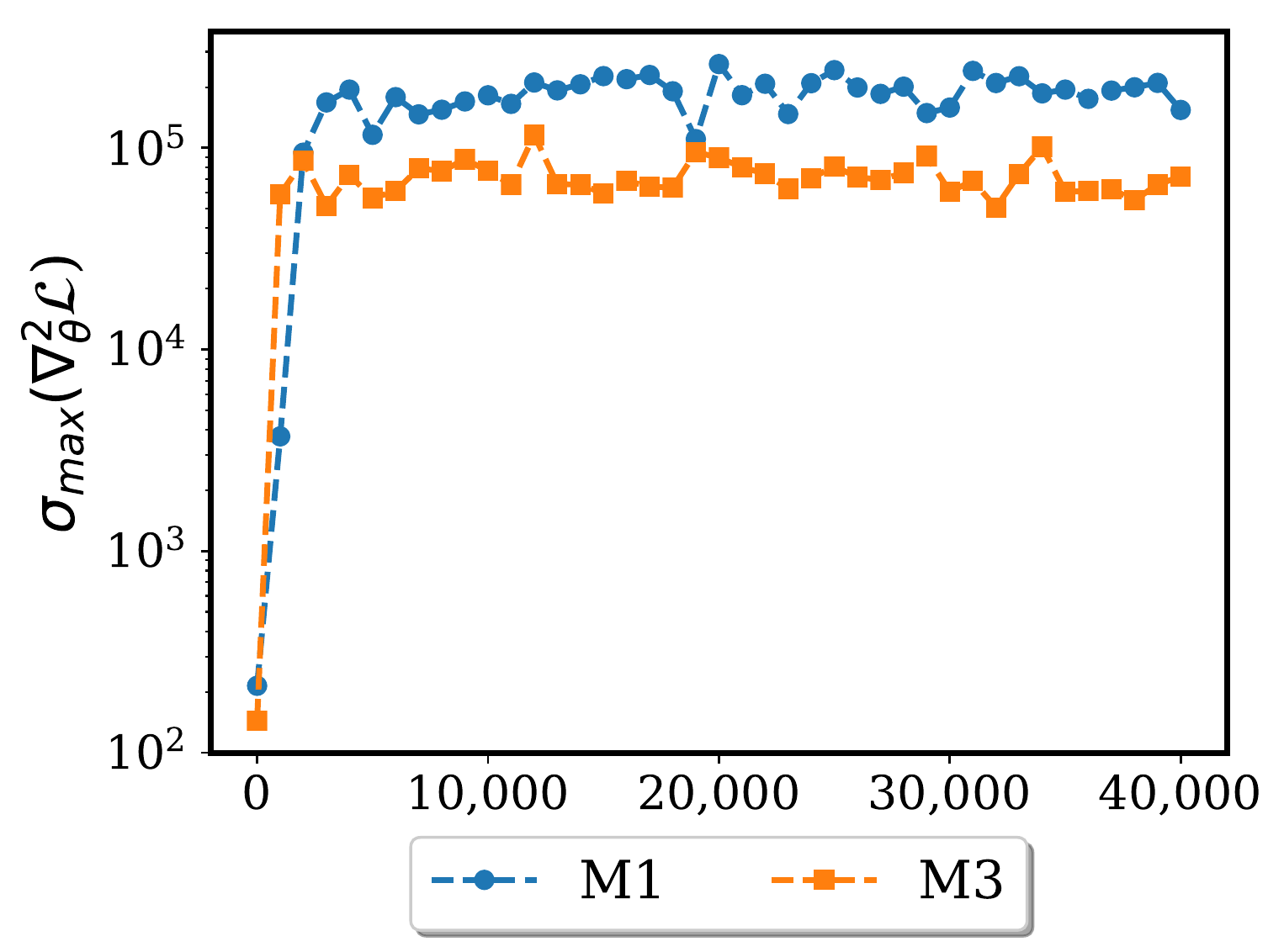}
    \caption{{\em An improved fully-connected architecture for physics-informed neural networks:} Largest eigenvalue of the Hessian  $\nabla_{\theta}^2\mathcal{L}(\theta)$ during the training of two different physics-informed neural network models for approximating the solution to the two-dimensional Helmholtz problem (see equation \ref{eq:Helmholtz}) with $a_1=1$, $a_2=4$. Model M1 is a conventional dense, 4-layer deep architecture, while model M3 corresponds to the proposed improved architecture also with a depth of 4 layers.}
    \label{fig:ADGM_stifness}
\end{figure}

Perhaps a more interesting fact, is that the choice of the neural architecture has also an influence on the stiffness of the gradient flow dynamics. To illustrate this point, we report the largest eigenvalue of the Hessian matrix $\sigma_{\max}(\nabla_{\theta}^2\mathcal{L}(\theta))$ for the Helmholtz benchmark discussed in section \ref{sec:gradient_pathologies} using a stiff parameter setting corresponding to $a_1=1$, $a_2=4$. Specifically, in figure \ref{fig:ADGM_stifness} we compare the evolution of  $\sigma_{\max}(\nabla_{\theta}^2\mathcal{L}(\theta))$ during model training for the conventional dense, 4-layer deep architecture employed in section \ref{sec:gradient_pathologies} (denoted as model M1), versus the proposed neural architecture also with a depth of 4 layers (denoted as model M3). Evidently, the proposed architecture yields a roughly 3x decrease in the magnitude of the leading Hessian eigenvalue, suggesting that tweaks in the neural network architecture could potentially stabilize and accelerate the training of physics-informed neural networks; a direction that will be explored in future work. Moreover, as reported in section \ref{sec:results}, the combination of the architecture and the learning rate annealing scheme put forth in section \ref{sec:adaptive} consistently improves the accuracy of physics-informed neural networks by a factor of 50-100x across a range of benchmark problems in computational physics.

\section{Results}
\label{sec:results}

In this section we provide a collection of comprehensive numerical studies that aim to assess the performance or the proposed methodologies against the current state-of-the-art in using fully-connected deep neural network models for inferring the solution of PDEs. Throughout all benchmarks we will employ and compare the performance of four different approaches as summarized in table \ref{tab:models}. In all cases we employ hyperbolic tangent activation functions, a mini-batch size of 128 data-points, and train the networks using single precision arithmetic via stochastic gradient descent using the Adam optimizer with default settings \cite{kingma2014adam}. Moreover, all networks are initialized using the Glorot scheme \cite{glorot2010understanding}, and no additional regularization techniques are employed (e.g., dropout, $L^1$/$L^2$ penalties, etc.). Models M2 and M4 use the proposed learning rate annealing algorithm update their loss weights $\lambda_i$ every 10 iterations of stochastic gradient descent (see equations \ref{eq:lambda_hat_update} and \ref{eq:lambda_update}). All algorithms are implemented in Tensorflow \cite{abadi2016tensorflow} and the reported run-times are obtained on a Lenovo X1 Carbon ThinkPad laptop with an Intel Core i5 2.3GHz processor and 8Gb of RAM memory. The code and data accompanying this manuscript is publicly available at \url{https://github.com/PredictiveIntelligenceLab/GradientPathologiesPINNs}.

\begin{table}
\centering
\begin{tabular}{|c|p{10cm}|}
\hline
Model & Description \\ \hline
M1  & Original formulation of physics-informed neural networks as put forth in Raissi {\it et. al.} \cite{raissi2019physics}          \\ \hline
M2  & Physics-informed neural networks with the proposed learning rate annealing algorithm (see section \ref{sec:adaptive})      \\ \hline
M3  & Physics-informed neural networks with the proposed improved fully-connected architecture (see section \ref{sec:improved_FC})   \\\hline
M4  & Physics-informed neural networks combining  the proposed learning rate annealing algorithm and fully-connected architecture        \\ \hline
\end{tabular}
\caption{Models considered in the numerical studies presented in section \ref{sec:results}.}
\label{tab:models}
\end{table}

\subsection{Helmholtz Equation}
First, let us revisit the Helmholtz equation benchmark described in section \ref{sec:gradient_pathologies} (see equation \ref{eq:Helmholtz}). This equation is closely related to many problems in natural and engineering sciences, such as wave propagation in acoustic, elastic and electromagnetic media \cite{nedelec2001acoustic}. It also routinely arises in conventional numerical discretization methods for PDEs such as finite element and spectral methods \cite{zienkiewicz1977finite, canuto2006spectral}. Our goal here is to use this canonical benchmark problem to systematically analyze the performance of the different models listed in table \ref{tab:models}, and quantify their predictive accuracy for different choices of the underlying neural network architecture. Specifically, we have varied the number of hidden layers and the number of neurons per layer that define each model's architecture and report the resulting relative $L^2$ error for the predicted solution after $40,000$ Adam iterations. 

Table \ref{tab:Helmholtz} summarizes our results, averaged over 10 independent trials with randomized weight initializations using the Glorot scheme \cite{glorot2010understanding}. Evidently, the original formulation of Raissi {\it et. al.} \cite{raissi2019physics} (model M1) is very sensitive to the choice of neural architecture and fails to attain a stable prediction accuracy, yielding errors ranging between $8$-$31\%$ in the relative $L^2$ norm. In contrast, the proposed models (models M2-M4) appear to be very robust with respect to network architectures and show a consistent trend in improving the prediction accuracy as the number of hidden layers and neural units is increased. We also observe that, by introducing a small number of additional weights and biases, the improved fully-connected architecture corresponding to model M3 can consistently yield better prediction accuracy than the conventional fully-connected architectures used in the original work of Raissi {\it et. al.} \cite{raissi2019physics} (model M1). This indicates that the proposed architecture seems to have better ability to represent complicated functions than conventional fully-connected neural networks, as well as it may lead to more stable gradient descent dynamics as discussed in section \ref{sec:improved_FC}. Last but not least, the combination of the proposed improved architecture with the adaptive learning rate algorithm of section \ref{sec:adaptive} (model M4) uniformly leads to the most accurate results we have obtained for this problem (relative $L^2$ errors ranging between $0.12$-$0.25\%$).


\begin{table}[h]
\centering
\begin{tabular}{|c|c|c|c|c|}
\hline
Architecture                & M1       & M2       & M3       & M4       \\ \hline
30 units / 3 hidden layers  & 2.44e-01 & 5.39e-02 & 5.31e-02 & 1.07e-02 \\ \hline
50 units / 3 hidden layers  & 1.06e-01 & 1.12e-02 & 2.46e-02 & 3.87e-03 \\ \hline
100 units / 3 hidden layers & 9.07e-02 & 4.84e-03 & 1.17e-02 & 2.71e-03 \\ \hline
30 units / 5 hidden layers  & 2.47e-01 & 2.74e-02 & 4.12e-02 & 3.49e-03 \\ \hline
50 units / 5 hidden layers  & 1.40e-01 & 7.43e-03 & 1.97e-02 & 2.54e-03 \\ \hline
100 units / 5 hidden layers & 1.15e-01 & 5.37e-03 & 1.08e-02 & 1.63e-03 \\ \hline
30 units / 7 hidden layers  & 3.10e-01 & 2.67e-02 & 3.17e-02 & 3.59e-03 \\ \hline
50 units / 7 hidden layers  & 1.98e-01 & 7.33e-03 & 2.37e-02 & 2.04e-03 \\ \hline
100 units / 7 hidden layers & 8.14e-02 & 4.52e-03 & 9.36e-03 & 1.49e-03 \\ \hline
\end{tabular}
\caption{{\em Helmholtz equation:} Relative $L^2$ error between the predicted and the exact solution $u(x,y)$ for the different methods summarized in table \ref{tab:models}, and for different neural architectures obtained by varying the number of hidden layers and different number of neurons per layer.}
\label{tab:Helmholtz}
\end{table}


\subsection{Klein-Gordon equation }
To emphasize the ability of the proposed methods to handle nonlinear and time-dependent problems, leading to composite loss functions with multiple competing objectives, let us consider the time-dependent Klein-Gordon equation. The Klein–Gordon equation is a non-linear equation, which plays a significant role in many scientific applications such as solid-state physics, nonlinear optics, and quantum physics \cite{bjorken1965relativistic}. The initial-boundary value problem in one spatial dimension takes the form
\begin{align}
\label{eq:klein-gordon}
    &u_{tt} + \alpha u_{xx} + \beta u + \gamma u^k = f(x,t), &&   (x, t) \in \Omega \times [0, T]  \\
    &u(x,0) = g_1(x), &&  x \in \Omega \\
    &u_t(x,0) = g_2(x), && x \in \Omega \\
    &u(x, t) = h(x, t), && (x, t) \in \partial \Omega \times [0, T]
\end{align}
where $\alpha, \beta, \gamma$ and $k$ are known constants, $k =2$ when we have a quadratic non-linearity and $k =3$ when we have a cubic non-linearity. The functions $f(x,t), g_1(x), g_2(x)$ and $h(x)$ are considered to be known, and the function $u(x,t)$ represents the latent solution to this problem. Here we choose $\Omega = [0,1] \times [0, 1], T=1, \alpha =-1, \beta = 0, \gamma =1, k=3$, and the initial conditions satisfying  $g_1(x) = g_2(x) = 0$ for all $x \in \Omega$. To assess the accuracy of our models we will use a fabricated solution 
\begin{align}
\label{eq:gordon-klein_exact}
    u(x,t) = x \cos(5 \pi t) + (xt)^3,
\end{align}
and correspondingly derive a consistent forcing term $f(x,t)$ using equation \ref{eq:klein-gordon}, while the Dirichlet boundary condition term $h(x, y)$ imposed on the spatial domain boundaries $\partial \Omega$ is directly extracted from the fabricated  solution of equation \ref{eq:gordon-klein_exact}. 

A physics-informed neural network model $f_{\theta}(x,t)$ can now be trained to approximate the latent solution $u(x,t)$ by formulating the following composite loss
\begin{align}
    \label{eq: loss_Klein_Gordon}
    \mathcal{L}(\theta) =   \mathcal{L}_r(\theta) + \lambda_{u_b}  \mathcal{L}_{u_b}(\theta) +  \lambda_{u_0} \mathcal{L}_{u_0}(\theta)
\end{align}
where $\mathcal{L}_r(\theta), \mathcal{L}_{u_b}(\theta),\mathcal{L}_{u_0}(\theta)$ are defined in equations \ref{eq: loss_r}, \ref{eq: loss_ub}, and \ref{eq: loss_u0}.  Notice that if $\lambda_{u_b} = \lambda_{u_0} = 1$, then the loss function above gives rise to the original physics-informed neural network formulation of Raissi {\it et. al.} \cite{raissi2019physics} (model M1). 

\begin{figure}
\centering
\includegraphics[width = 0.8\linewidth]{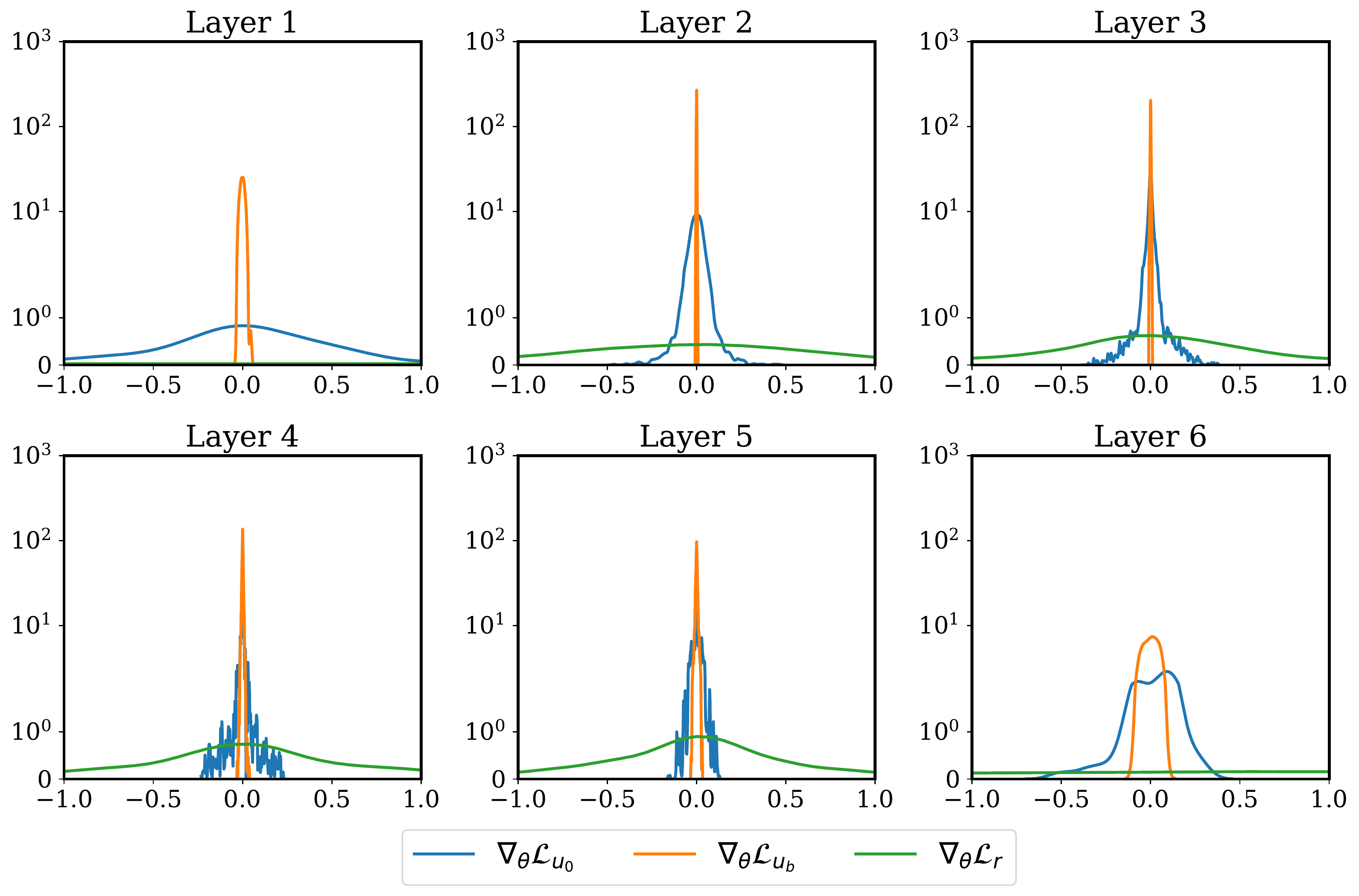}
\caption{{\em Klein-Gordon equation:} Histograms of back-propagated gradients for model M1 after 40,000 iterations of stochastic gradient descent using the Adam optimizer.}
\label{fig:Klein_Gordon_gradients_original}
\end{figure}

Given this problem setup, let us now investigate the distribution of back-propagated gradients during the training of a 5-layer deep fully-connected network with 50 neurons per layer using the the original formulation of Raissi {\it et. al.} \cite{raissi2019physics} (model M1). The results summarized in figure \ref{fig:Klein_Gordon_gradients_original} indicate that $\nabla_\theta \mathcal{L}_{u_b}$ and $\nabla_\theta \mathcal{L}_{u_0}$ accumulate at the origin around which they form two sharp peaks, while $\nabla_\theta \mathcal{L}_{r}$ keeps flat. This implies that the gradients of $\mathcal{L}_{u_b}$ and $\mathcal{L}_{u_r}$ nearly diminish in comparison to the gradients of $\mathcal{L}_{r}$. This is a clear manifestation of the imbalanced gradient pathology described in section \ref{sec:gradient_pathologies}, which we believe is the leading mode of failure for conventional physics-informed neural network models. Consequently, model M1 fails to accurately fit the initial and boundary conditions data, and therefore, as expected, yields a poor prediction for the target solution $u(x,t)$ with a relative $L^2$ error of $17.9\%$. From our experience, this behavior is extremely common in using conventional physics-informed neural network models \cite{raissi2019physics} for solving PDE systems that have sufficiently complex dynamics leading to solutions with non-trivial behavior (e.g., directional anisotropy, multi-scale features, etc.).

\begin{figure}
\centering
\includegraphics[width = 0.8\linewidth]{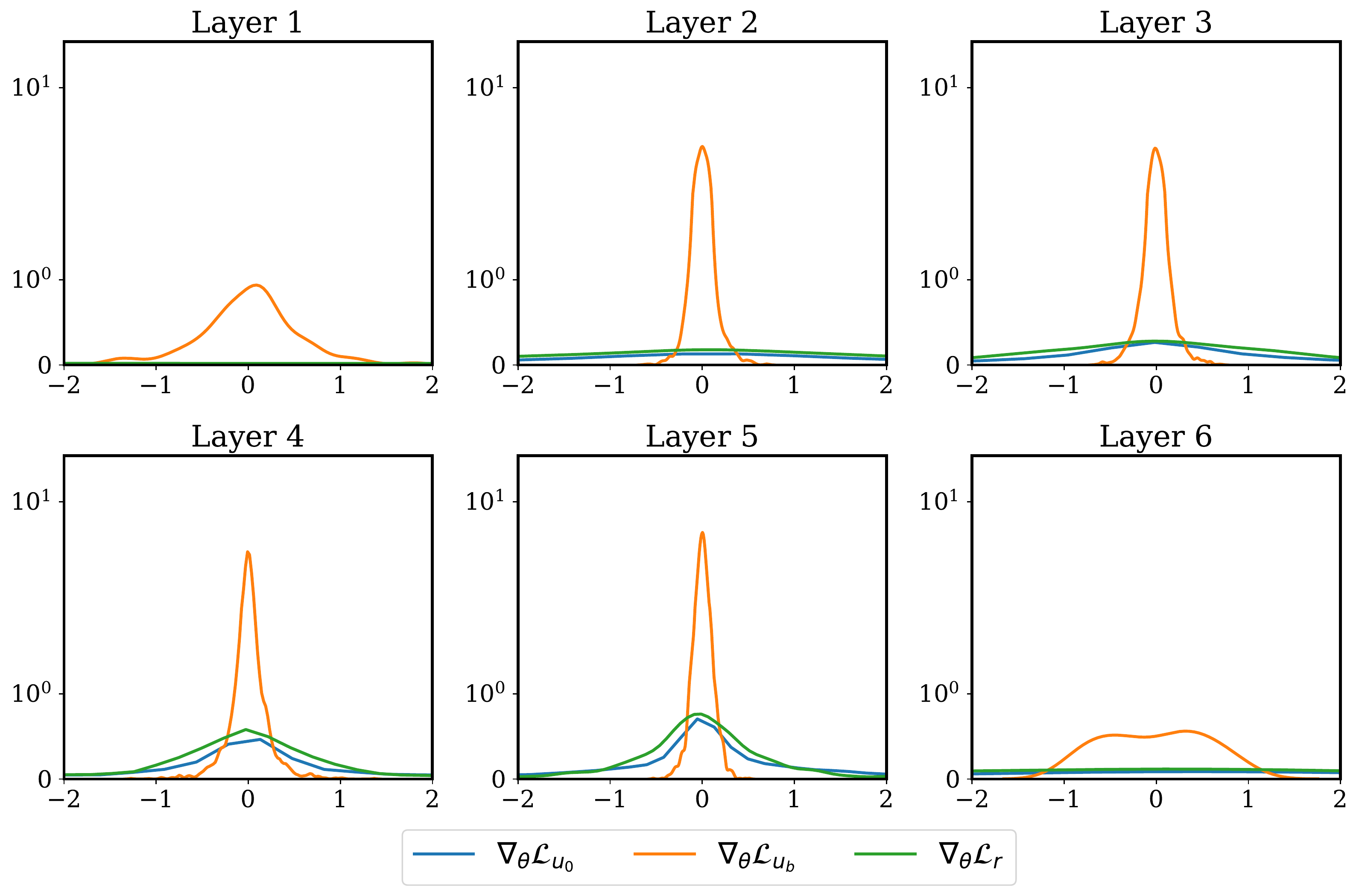}
\caption{{\em Klein-Gordon equation:} Back-propagated gradients for model M2 after 40,000 iterations of stochastic gradient descent using the Adam optimizer.}
\label{fig:Klein_Gordon_gradients_adaptive}
\end{figure}

It is now natural to ask whether the proposed methods can effectively mitigate this gradient imbalance pathology and lead to accurate and robust predictions. To this end, in figure \ref{fig:Klein_Gordon_gradients_adaptive} we present the distribution of back-propagated gradients for $\nabla_\theta \mathcal{L}_{r}$, $\nabla_\theta \mathcal{L}_{u_b}$, $\nabla_\theta \mathcal{L}_{u_0}$ at each hidden layer of the same neural architecture used above, albeit using the proposed learning rate annealing algorithm (model M2) for adaptively choosing the constants $\lambda_{u_b}$ and $\lambda_{u_0}$ during model training. We observe that all gradient distributions do not sharply peak around zero values indicating the fact that a healthy gradient signal is back-propagated through the network from all the contributing terms in the loss function.

A comparative study on the performance of all models (M1-M4) is summarized in table \ref{tab:Klein_Gordon}. Notice that here our goal is not to perform an exhaustive hyper-parameter search to find the best performing model, rather our goal is to present indicative results for the performance of each method (models M1-M4). To do so, we consider a fixed neural architecture with 5 hidden layers and 50 neurons and report the relative $L^2$ prediction error of each method after $40,000$ iterations of training, along with the total wall clock time required to complete each simulation. It is clear that the proposed methodologies consistently outperform the original implementation of Raissi {\it et. al.} \cite{raissi2019physics}, while the combined use of the proposed learning rate annealing algorithm with the improved fully-connected architecture (model M4) lead to the best results, albeit at a higher computational cost.


\begin{table}
    \centering
    \begin{tabular}{|c|c|c|c|c|}
    \hline
    Method         & M1 & M2 & M3     & M4 \\ \hline
    Relative $L^2$ error       & 1.79e-01 & 1.06e-02 & 1.97e-02 & 2.81e-03      \\ \hline
    Training time (sec)       & 200 & 200 & 760  &  1040  \\ \hline
\end{tabular}
\caption{{\em Klein-Gordon equation:} Relative $L^2$ error between the predicted and the exact solution $u(t, x)$ for the different methods summarized in table \ref{tab:models}. Here the network architecture is fixed to 5 fully-connected layers with 50 neurons per layer. The reported CPU time corresponds to the total wall-clock time to train each network for 40,000 iterations of stochastic gradient descent on a Lenovo X1 Carbon ThinkPad laptop with an Intel Core i5 2.3GHz processor and 8Gb of RAM memory.}
\label{tab:Klein_Gordon}
\end{table}

\begin{figure}
    \centering
    \includegraphics[width = \linewidth]{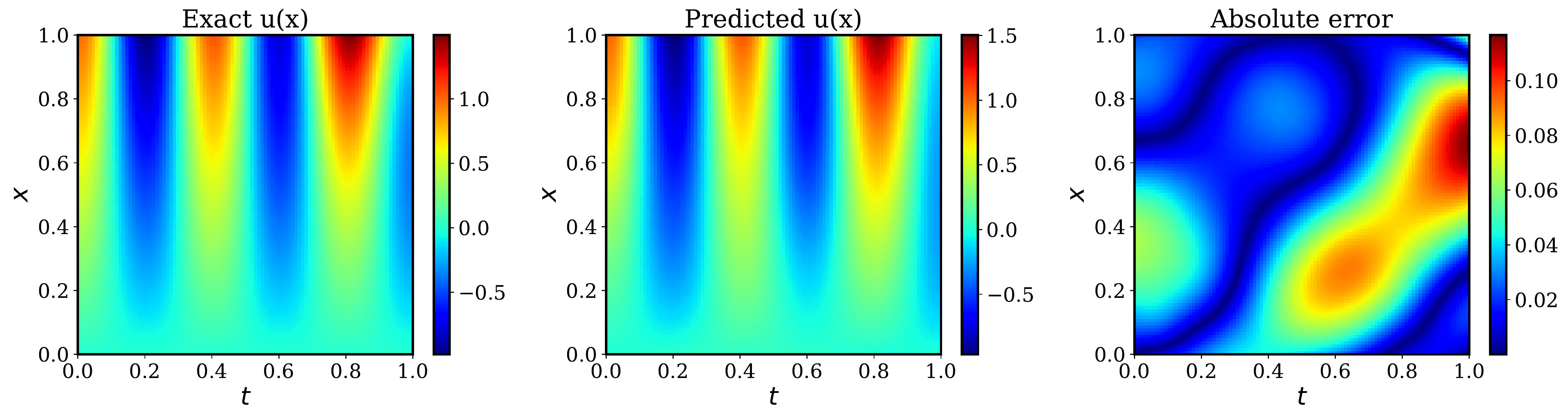}
    \caption{{\em Klein Gordon equation:} Predicted solution of model M1, a conventional physics-informed neural network \cite{raissi2019physics} with 5 hidden layers and 50 neurons in each layer.The relative $L^2$ error of the prediction solution is  1.79e-01  after 40,000 iterations of stochastic gradient descent using the Adam optimizer.}
    \label{fig:Klein_Gordon_prediction_original}
\end{figure}

\begin{figure}
    \centering
    \includegraphics[width = \linewidth]{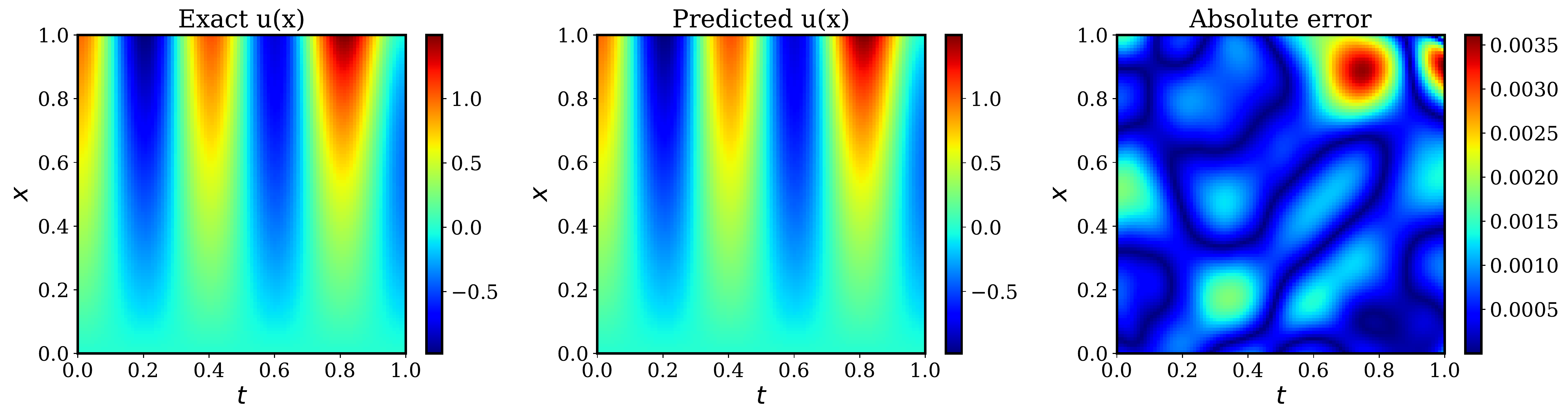}
    \caption{{\em Klein Gordon equation:} Predicted solution of model M4, a physics-informed neural network \cite{raissi2019physics} with 5 hidden layers and 50 neurons in each layer trained using the proposed learning rate annealing algorithm (see section \ref{sec:adaptive}), and the improved architecture described in \ref{sec:improved_FC}. The relative $L^2$ error of the prediction solution is  2.22e-03  after 40,000 iterations of stochastic gradient descent using the Adam optimizer.}
    \label{fig:Klein_Gordon_prediction_adaptive_ADGM}
\end{figure}

Figures \ref{fig:Klein_Gordon_prediction_original} and \ref{fig:Klein_Gordon_prediction_adaptive_ADGM} provide a more detailed visual assessment of the predictive accuracy for models M1 and M4. As we expected, in figure \ref{fig:Klein_Gordon_prediction_original}, model M1 fails to obtain accurate predictive solutions and has increased absolute error up to about $0.2$ as time $t$ increases. In figure \ref{fig:Klein_Gordon_prediction_adaptive_ADGM}, we present the solution obtained model M4 that combines the proposed learning rate annealing algorithm (see section \ref{sec:adaptive}) and improved neural architecture put forth in section \ref{sec:improved_FC}. It can be observed that the absolute error takes a maximum value of $0.004$ and the relative $L^2$ prediction error is $0.28\%$, which is about two orders of magnitude lower than the one obtained using model M1, and one order of magnitude lower that the predictive error of models M2 and M3. Finally, figure \ref{fig:Klein_Gordon_gradients_adaptive} depicts the  evolution of the constants $\lambda_{u_{b}}$ and $\lambda_{u_0}$ used to scale the boundary and initial conditions loss, $\mathcal{L}_{u_{b}}(\theta)$ and $\mathcal{L}_{u_{0}}(\theta)$, respectively (see equation \ref{eq: loss_Klein_Gordon}), during the training of model M2 using algorithm \ref{alg:learning_rate_annealing}.
 
\begin{figure}
    \centering
    \includegraphics[width = 0.5\textwidth]{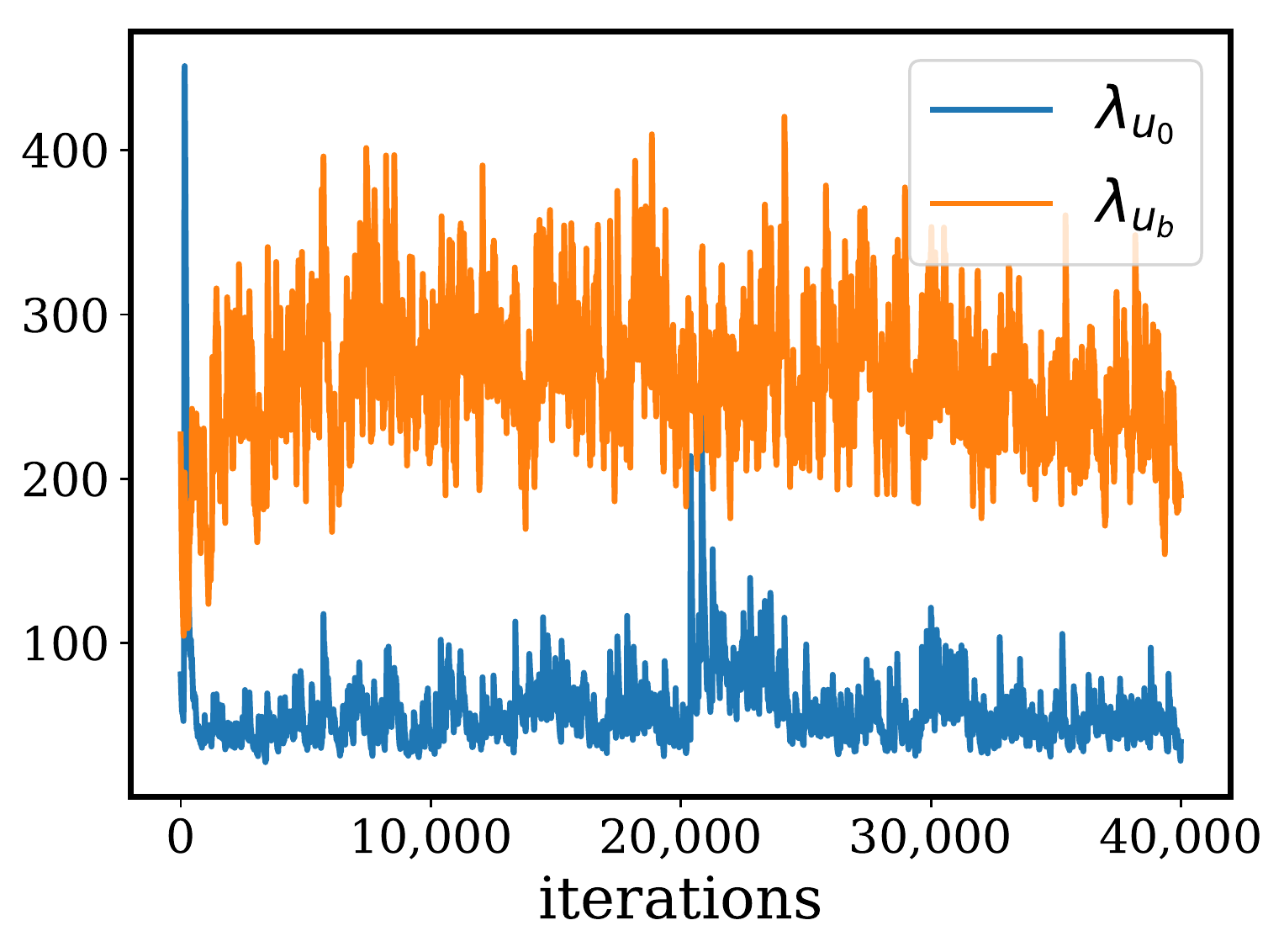}
    \caption{{\em Klein Gordon equation:} Evolution of the constants $\lambda_{u_{b}}$ and $\lambda_{u_0}$ used to scale the boundary and initial conditions loss, $\mathcal{L}_{u_{b}}(\theta)$ and $\mathcal{L}_{u_{0}}(\theta)$, respectively (see equation \ref{eq: loss_Klein_Gordon}), during the training of model M2 using algorithm \ref{alg:learning_rate_annealing}.}
    \label{fig:Klein_Gordon_adaptive_constant}
\end{figure}

\subsection{Flow in a lid-driven cavity}

In our last example, we present a study that highlights a few important remarks on how a problem's formulation can be as crucial as the algorithm used to solve it. To this end, let us consider a classical benchmark problem in computational fluid dynamics, the steady-state flow in a two-dimensional lid-driven cavity. The system is governed by the incompressible Navier-Stokes equations which can be written in non-dimensional form as
\begin{align}
    \label{eq: NS}
    &\bm{u}\cdot\nabla{\bm{u}} + \nabla p - \frac{1}{Re}\Delta \bm{u} = 0 &&\text{in } \Omega \\
    &\nabla \cdot \bm{u} = 0  &&\text{in } \Omega \\
    &\bm{u}(\bm{x}) = (1, 0)    &&\text{on } \Gamma_1 \\
    &\bm{u}(\bm{x}) = (0, 0)    &&\text{on } \Gamma_0
\end{align}
where $\bm{u}(\bm{x}) = (u(\bm{x}), v(\bm{x}))$ is a velocity vector field, $p$ is a scalar pressure field,  and $\bm{x}=(x,y)\in\Omega = (0,1) \times (0, 1)$ with $\Omega$ being a two-dimensional square cavity. Moreover, $\Gamma_1$ is the top boundary of the cavity,  $\Gamma_0$ denotes the other three sides, and $Re$ is the Reynolds number of the flow. In this example, we chosen a relatively simple case corresponding to a Reynolds number of $Re =100$, for which it is well understood that the flow quickly converges to an incompressible steady-state solution \cite{bruneau20062d}. Our goal here is to train a physics-informed neural network to predict the latent velocity and pressure fields. In order to assess the accuracy of our predictions, we have also simulated this system to obtain a reference solution using conventional finite difference methods, there-by creating a high-resolution validation data set $\{\bm{x^i}, \bm{u}^i\}_{i=1}^N$ (see details in Appendix \ref{sec:appendix_FDM}). In the following subsections we will consider two different neural network representations to highlight a crucial aspect of simulating incompressible fluid flows with physics-informed neural networks.

\subsection{Velocity-pressure representation}

A straightforward application of physics-informed neural networks for approximating the solution of equation \ref{eq: NS} would introduce a neural network 
representation with parameters $\theta$ that aims to learn how to map spatial coordinates $(x,y)$ to the latent solution functions $u(x,y)$, $v(x,y)$ and $p(x,y)$, i.e.,
$$[x, y]\xmapsto[ ]{f_{\theta}}[u(x, y), v(x, y), p(x, y)]$$
Using this representation one can then introduce the residuals of the momentum and continuity equations as
\begin{align}
\label{eq:NS_uvp_residuals}
r_{\theta}^{u}(x,y) &:= u\frac{\partial u}{\partial x} + v\frac{\partial u}{\partial y} + \frac{\partial p}{\partial x} - \frac{1}{Re}(\frac{\partial^2 u}{\partial x^2} + \frac{\partial^2 u}{\partial y^2}) \\
r_{\theta}^{v}(x,y) &:= u\frac{\partial v}{\partial x} + v\frac{\partial v}{\partial y} + \frac{\partial p}{\partial y} - \frac{1}{Re}(\frac{\partial^2 v}{\partial x^2} + \frac{\partial^2 v}{\partial y^2}) \\
r_{\theta}^{c}(x,y) &= \frac{\partial u}{\partial x} + \frac{\partial v}{\partial y},
\end{align}
where $(u,v,p) = f_{\theta}(x,y)$ are the outputs of the chosen neural network representation, and all required gradients can be readily computed using automatic differentiation \cite{baydin2018automatic}. Given these residuals, along with a set of appropriate boundary conditions for equation \ref{eq: NS} we can now formulate a loss function for training a physics-informed neural network as
\begin{align}
\label{eq:NS_uvp_loss}
\mathcal{L}(\theta) = \mathcal{L}_{r_u}(\theta) + \mathcal{L}_{r_v}(\theta) + \mathcal{L}_{r_c}(\theta) + \mathcal{L}_{u_b}(\theta) + \mathcal{L}_{v_b}(\theta),
\end{align}
with 
\begin{align}
\label{eq:NS_uvp_mse}
    &\mathcal{L}_{r_u}(\theta) = \frac{1}{N_r} \sum_{i=1}^{N_r} [r_{\theta}^{u}(x_r^{i},y_r^{i})]^2 \\
    &\mathcal{L}_{r_v}(\theta) = \frac{1}{N_r} \sum_{i=1}^{N_r} [r_{\theta}^{v}(x_r^{i},y_r^{i})]^2 \\
    &\mathcal{L}_{r_c}(\theta) = \frac{1}{N_r} \sum_{i=1}^{N_r} [r_{\theta}^{c}(x_r^{i},y_r^{i})]^2 \\
    &\mathcal{L}_{u_b}(\theta) = \frac{1}{N_b} \sum_{i=1}^{N_b}[u(x_b^{i},y_b^{i}) - u_b^i]^2, \\
    &\mathcal{L}_{v_b}(\theta) = \frac{1}{N_b} \sum_{i=1}^{N_b}[v(x_b^{i},y_b^{i}) - v_b^i]^2
\end{align}
where $\{(x_r^i, y_r^i)\}_{i=1}^{N_r}$ is a set of collocation points in which we aim to minimize the PDE residual, while $\{(x_b^i, y_b^i), u_b^i\}_{i=1}^{N_b}$ and $\{(x_b^i, y_b^i), v_b^i\}_{i=1}^{N_b}$ denote the boundary data for the two velocity components at the domain boundaries $\Gamma_0$ and $\Gamma_1$, respectively. 
Taken together, these terms aim to constrain the neural network approximation to satisfy the Navier-Stokes system and boundary conditions prescribed in equation \ref{eq: NS}. In practice, this composite loss is minimized using stochastic gradient descent, where a different set of collocation and boundary points are randomly sampled at each gradient descent iteration.

\begin{figure}
    \centering
     \begin{subfigure}[b]{\textwidth}
         \centering
         \includegraphics[width=\textwidth]{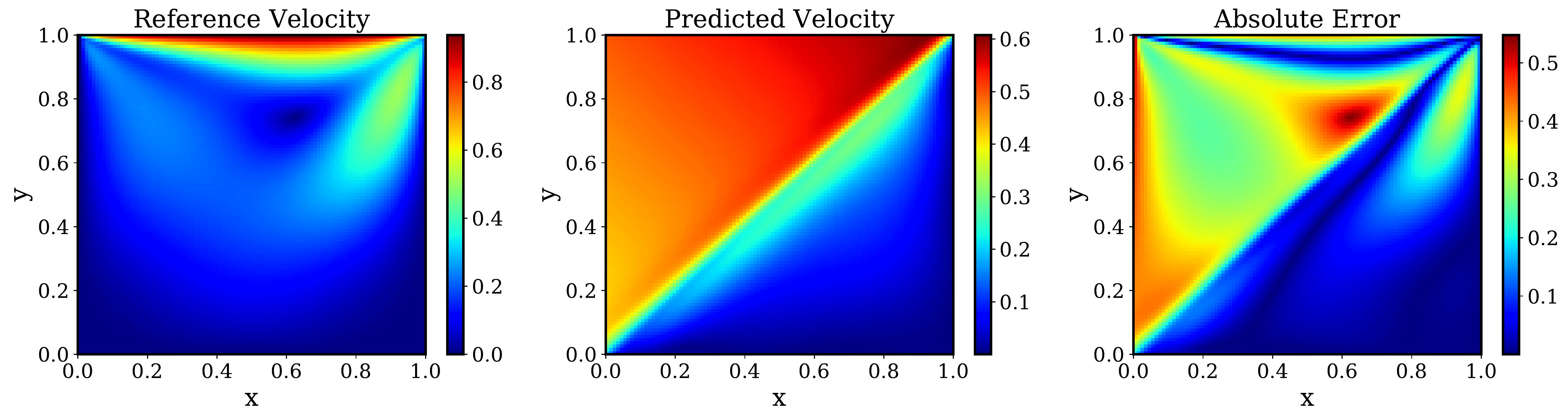}
         \caption{Model M1 (relative $L^2$ prediction error: 7.84e-01).}
         \label{fig: uvp_M1_velocity}
     \end{subfigure}
     \begin{subfigure}[b]{\textwidth}
         \centering
         \includegraphics[width=\textwidth]{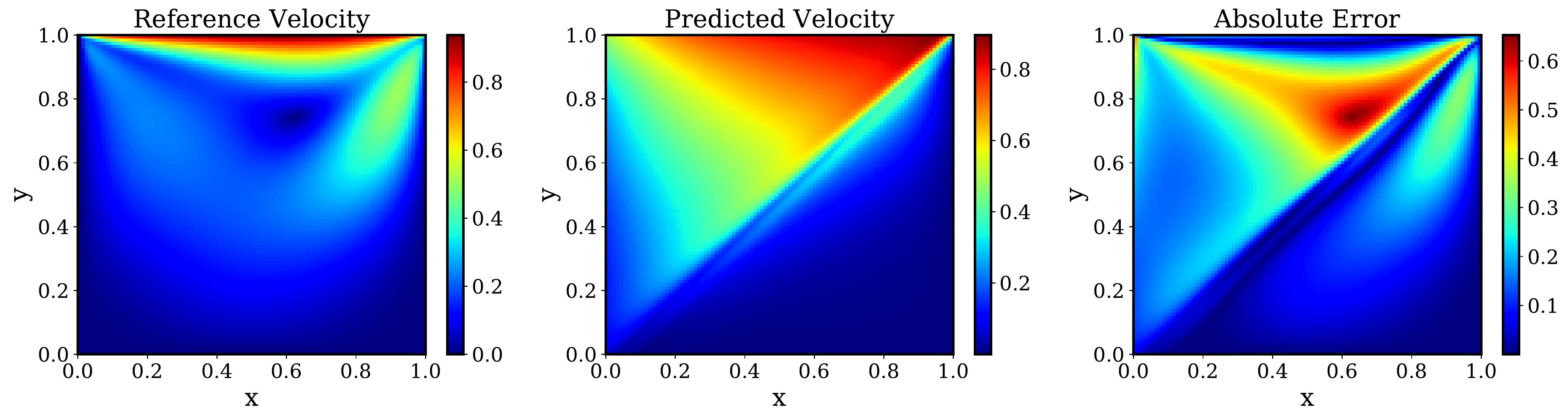}
         \caption{Model M2 (relative $L^2$ prediction error: 7.48e-01).}
         \label{fig: uvp_M2_velocity}
     \end{subfigure}
     \begin{subfigure}[b]{\textwidth}
         \centering
         \includegraphics[width=\textwidth]{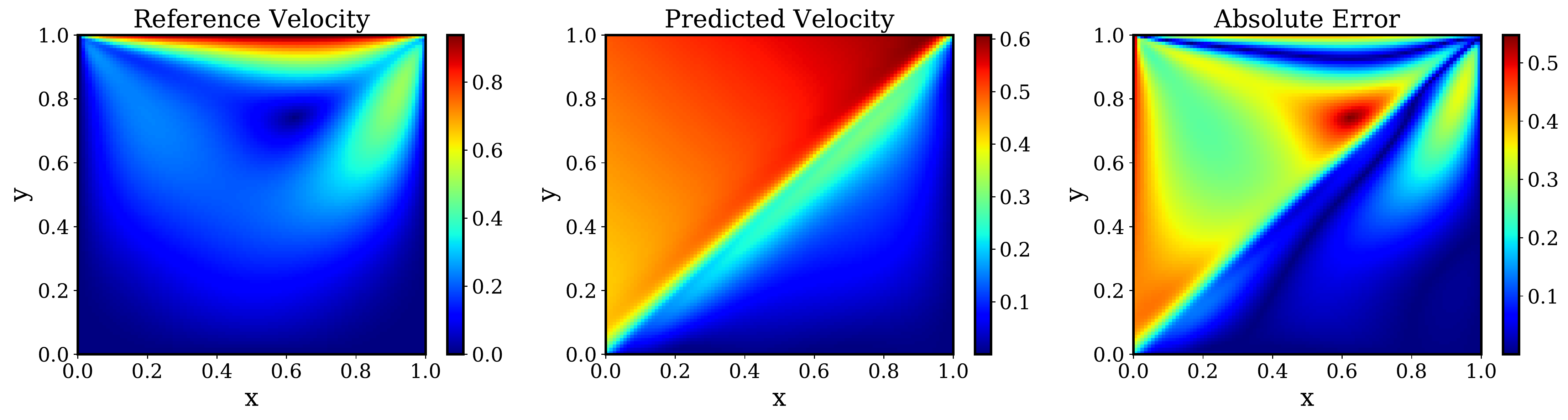}
         \caption{Model M3 (relative $L^2$ prediction error: 7.48e-01).}
         \label{fig: uvp_M3_velocity}
     \end{subfigure}
     \begin{subfigure}[b]{\textwidth}
         \centering
         \includegraphics[width=\textwidth]{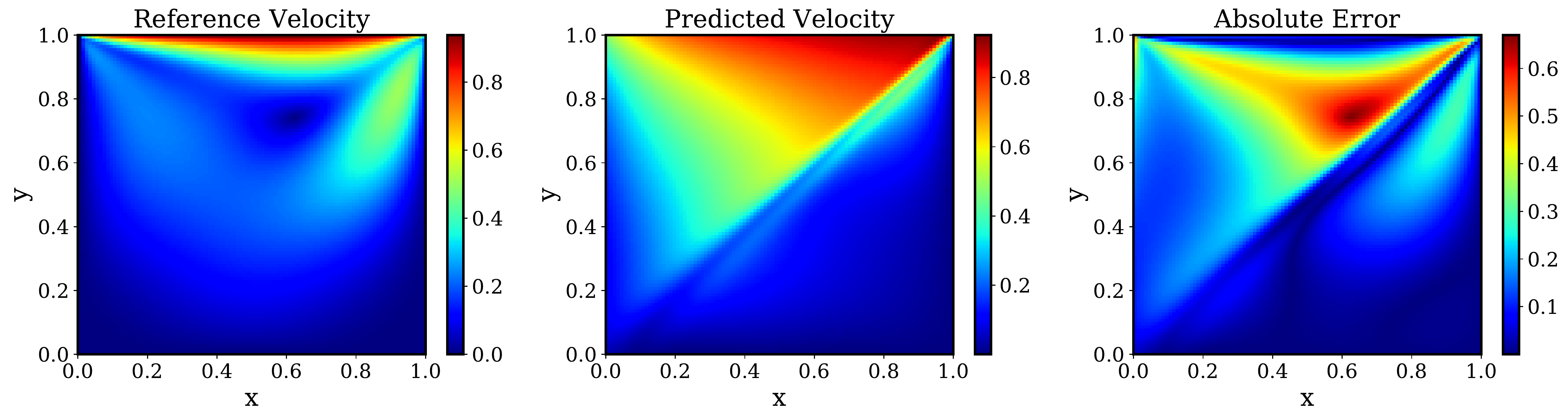}
         \caption{Model M4 (relative $L^2$ prediction error: 7.53e-01).}
         \label{fig: uvp_M4_velocity}
     \end{subfigure}
    \caption{{\em Flow in a lid-driven cavity, velocity-pressure representation:} Reference solution using a conventional finite difference solver, prediction of a 5-layer deep physics-informed neural network (models M1-M4, top to bottom), and absolute point-wise error.} 
    \label{fig:pred_velocity_uvp}
\end{figure}

Here, our goal is to learn the two velocity components $u(x,y)$ and $v(x,y)$, as well as the latent pressure field $p(x,y)$ by training a 5-layer deep neural network using the mean squared error loss of \ref{eq:NS_uvp_loss}. Each hidden layer contains 50 neurons and a hyperbolic tangent activation function. Figure \ref{fig:pred_velocity_uvp} summarizes the results of our experiment, which shows the magnitude of the predicted solution $|\bm u(x)| = \sqrt{u^2({\bm x}) + v^2({\bm x})}$ predicted by each model M1-M4, after $40,000$ stochastic gradient descent updates using the Adam optimizer \cite{kingma2014adam}. It is evident that none of the models is capable of producing a reasonable approximation. We postulate that this failure is related to the residual loss used to train the models. Specifically, it is inevitable that solutions inferred by minimizing equation \ref{eq:NS_uvp_loss} will not exactly satisfy the prescribed boundary conditions, nor the incompressibility constraint, potentially giving rise to issues of non-uniqueness. This fundamental limitation highlights the importance of adopting a consistent problem formulation, leading to effective objectives for training the physics-informed neural networks. As we will see in the next section, representing the latent velocity field as the gradient of a scalar potential can guarantee incompressible solutions and lead to stable training dynamics, as well as accurate predictions.

\subsection{Streamfunction-pressure representation}

Instead of directly representing the two velocity components $u(x,y)$ and $v(x,y)$, one can  use a neural network represent a scalar potential function $\psi(x,y)$ such that an incompressible velocity field can be directly computed as $(u,v) = (\partial \psi/\partial y, -\partial \psi/\partial x)$. Hence, we can introduce a neural network 
representation with parameters $\theta$ that aims to learn how to map spatial coordinates $(x,y)$ to the latent scalar functions $\psi(x,y)$, $p(x,y)$, i.e.,

$$[x, y]\xmapsto[ ]{f_{\theta}}[\psi(x, y), p(x, y)]$$

Then, we only need to consider the residuals for the momentum equations, as the incompressibility constraint is now exactly satisfied, i.e.
\begin{align}
\label{eq:NS_psip_residuals_u}
r_{\theta}^{u}(x,y) &:= u\frac{\partial u}{\partial x} + v\frac{\partial u}{\partial y} + \frac{\partial p}{\partial x} - \frac{1}{Re}(\frac{\partial^2 u}{\partial x^2} + \frac{\partial^2 u}{\partial y^2}) \\
r_{\theta}^{v}(x,y) &:= u\frac{\partial v}{\partial x} + v\frac{\partial v}{\partial y} + \frac{\partial p}{\partial y} - \frac{1}{Re}(\frac{\partial^2 v}{\partial x^2} + \frac{\partial^2 v}{\partial y^2}) \label{eq:NS_psip_residuals_v}
\end{align}
where $(u,v) = (\partial \psi/\partial y, -\partial \psi/\partial x)$, $(\psi,p) = f_{\theta}(x,y)$ are the outputs of the chosen neural network representation $f_{\theta}$, and all required gradients are computed using automatic differentiation \cite{baydin2018automatic}. Although we are now able to exactly satisfy the incompressibilty constraint, notice that this approach incurs a larger computational cost since third-order derivatives of the neural network are need to be computed. Despite this additional cost, our experience indicates that this modification is crucial for obtaining accurate and unique solutions for equation \ref{eq: NS} by minimizing a physics-informed loss of the form
\begin{align}
\label{eq:NS_psip_loss}
\mathcal{L}(\theta) = \mathcal{L}_{r_u}(\theta) + \mathcal{L}_{r_v}(\theta) + \mathcal{L}_{u_b}(\theta) + \mathcal{L}_{v_b}(\theta),
\end{align}
where 
\begin{align}
\label{eq:NS_psip_mse}
    &\mathcal{L}_{r_u}(\theta) = \frac{1}{N_r} \sum_{i=1}^{N_r} [r_{\theta}^{u}(x_r^{i},y_r^{i})]^2 \\
    &\mathcal{L}_{r_v}(\theta) = \frac{1}{N_r} \sum_{i=1}^{N_r} [r_{\theta}^{v}(x_r^{i},y_r^{i})]^2 \\
    &\mathcal{L}_{u_b}(\theta) = \frac{1}{N_b} \sum_{i=1}^{N_b}[u(x_b^{i},y_b^{i}) - u_b^i]^2, \\
    &\mathcal{L}_{v_b}(\theta) = \frac{1}{N_b} \sum_{i=1}^{N_b}[v(x_b^{i},y_b^{i}) - v_b^i]^2
\end{align}
where $\{(x_r^i, y_r^i)\}_{i=1}^{N_r}$ is a set of collocation points in which we aim to minimize the residual of equations \ref{eq:NS_psip_residuals_u} and \ref{eq:NS_psip_residuals_v}, while $\{(x_b^i, y_b^i), u_b^i\}_{i=1}^{N_b}$ and $\{(x_b^i, y_b^i), v_b^i\}_{i=1}^{N_b}$ denote the boundary data for the two velocity components at the domain boundaries $\Gamma_0$ and $\Gamma_1$, respectively. 

\begin{figure}
    \centering
     \begin{subfigure}[b]{\textwidth}
         \centering
         \includegraphics[width=\textwidth]{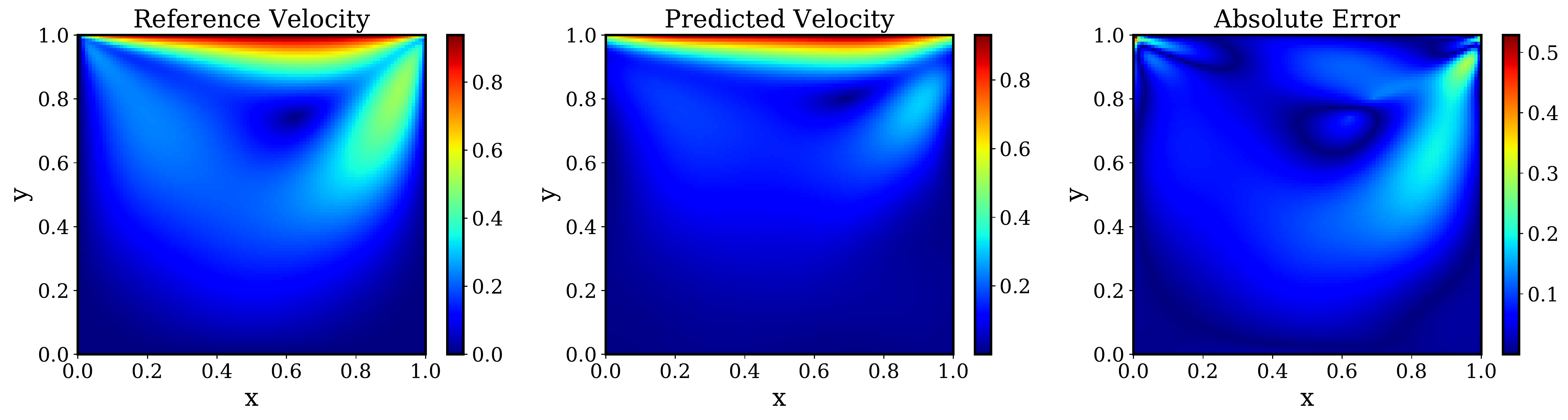}
         \caption{Model M1 (relative $L^2$ prediction error: 2.71e-01).}
         \label{fig: psi_p_M1_velocity}
     \end{subfigure}
     \begin{subfigure}[b]{\textwidth}
         \centering
         \includegraphics[width=\textwidth]{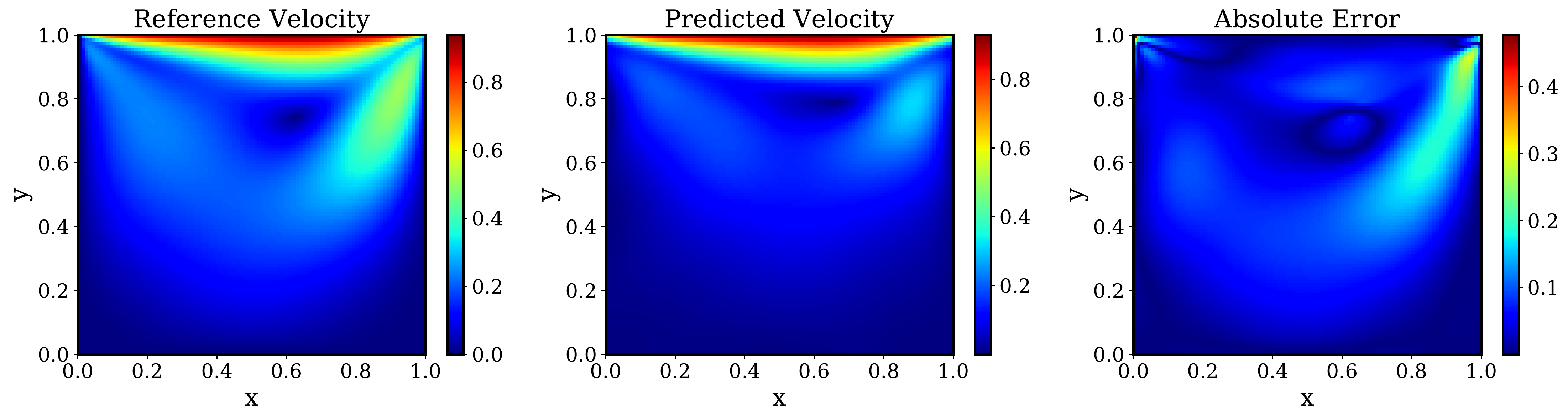}
         \caption{Model M2 (relative $L^2$ prediction error: 2.49e-01).}
         \label{fig: psi_p_M2_velocity}
     \end{subfigure}
     \begin{subfigure}[b]{\textwidth}
         \centering
         \includegraphics[width=\textwidth]{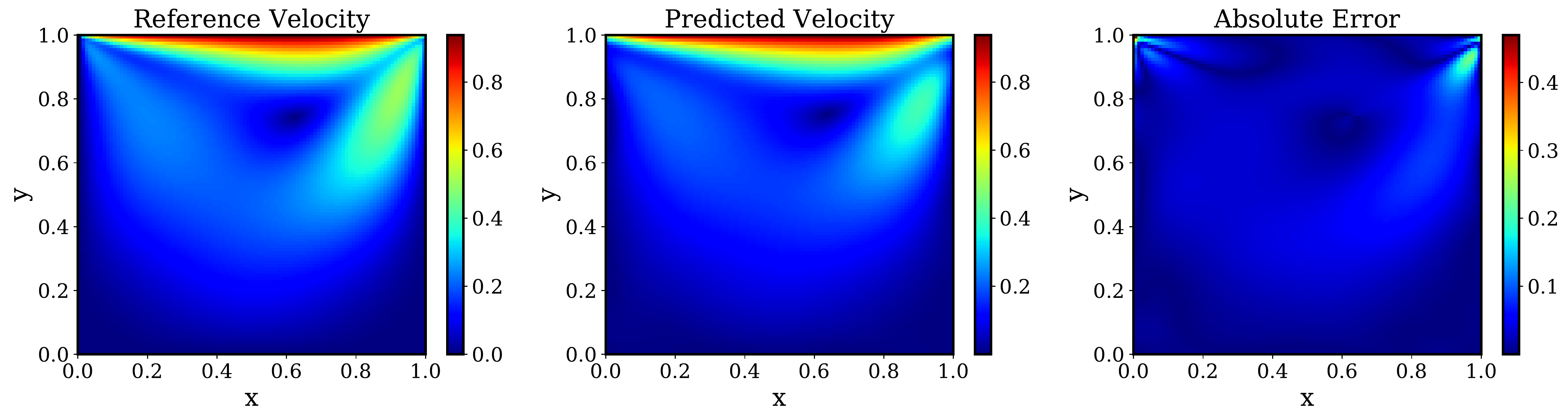}
         \caption{Model M3 (relative $L^2$ prediction error: 1.21e-01).}
         \label{fig: psi_p_M3_velocity}
     \end{subfigure}
     \begin{subfigure}[b]{\textwidth}
         \centering
         \includegraphics[width=\textwidth]{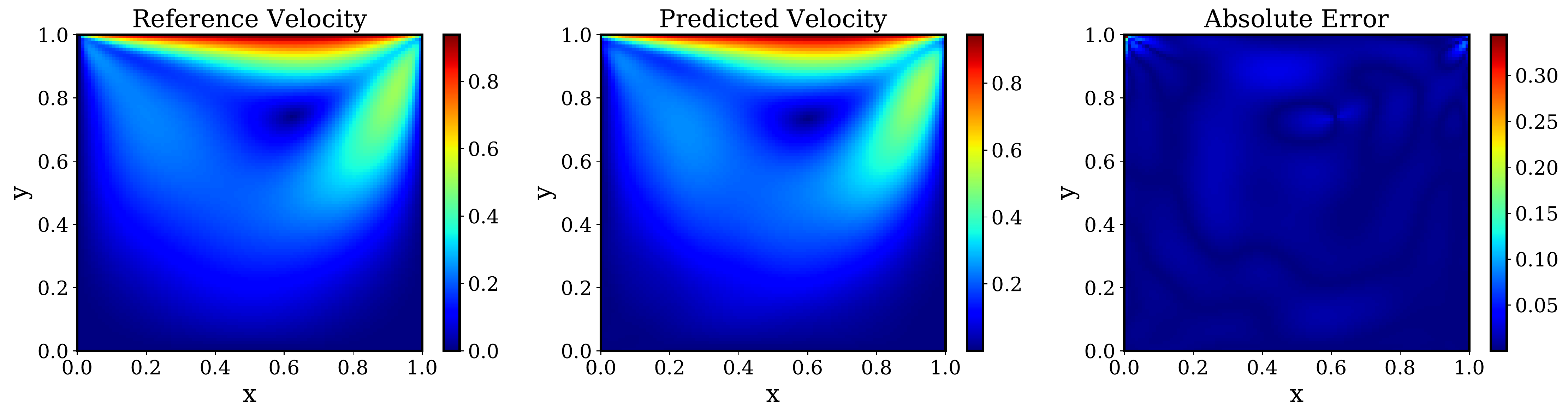}
         \caption{Model M4 (relative $L^2$ prediction error: 3.42e-02).}
         \label{fig: psi_p_M4_velocity}
     \end{subfigure}
    \caption{{\em Flow in a lid-driven cavity, streamfunction-pressure representation:} Reference solution using a conventional finite difference solver, prediction of a 5-layer deep physics-informed neural network (models M1-M4, top to bottom), and absolute point-wise error.} 
    \label{fig:pred_velocity_psip}
\end{figure}

Similarly, we still aim to learn the velocity components $u(x,y)$ and $v(x,y)$, as well as the latent pressure field $p(x,y)$. Now the velocity components can be obtained by taking derivatives of the stream function $\psi(x,y)$ with respect to the $x$ and $y$ coordinates using automatic differentiation \cite{baydin2018automatic}. We chose to jointly represent the latent function $[\psi(x, y), p(x, y)]$ using a 5-layer deep neural network with
50 neurons per hidden layer and a hyperbolic tangent activation function, trained using the mean squared error loss of equation \ref{eq:NS_psip_mse}. The results of this experiment are summarized in figure \ref{fig:pred_velocity_psip}. In particular, we present a comparison between the reference and the predicted solutions obtained using the different models M1 - M4. It can be observed that the velocity fields obtained using M2-M4 are in good agreement with the reference solution while M1 is not able to yield satisfactory prediction solution. As expected, model M4 yields the best predicted solution with an error of $3\%$ measured in the relative $L^2$-norm.


\section{Summary and Discussion}
\label{sec:discussion}

Despite recent success across a range of applications \cite{raissi2019physics, tartakovsky1808learning, kissas2020machine, tripathy2018deep, sun2019surrogate, raissi2018hidden}, physics-informed neural networks (PINNs) often struggle to accurately approximate the solution of partial differential equations. In this work we identify and analyze a fundamental mode of failure of physics-informed neural networks related to stiff  gradient flow dynamics that lead to unbalanced gradients during model training via back-propagation. 
To provide further insight, we quantify and analyze the stiffness of the gradient flow dynamics and elucidate the difficulties of training PINNs via gradient descent. This stiffness analysis not only plays a key role in understanding the gradient flow dynamics, but also motivates future research directions to consider more stable discretizations of the gradient flow system.
We mitigate this shortcoming by proposing a learning rate annealing algorithm that utilizes gradient statistics during model training to adaptive assign appropriate weights to different terms in a PINNs loss function. The new algorithm aims to address the unbalanced gradients pathology and  effectively lead to noticeable improvements in predictive accuracy. 
This development is not limited to PINNs, but can be straightforwardly generalized to other tasks involving the interplay of multiple objective functions that may lead to unbalanced gradients issue, e.g. multi-task learning \cite{chen2017gradnorm}.
Finally, we propose a novel neural network architecture that can enhance the accuracy and robustness of PINNs by decreasing the stiffness of gradient flow, indicating that specialized architectures can play a prominent role in the future success of PINNs models. Taken together, our developments provide new insights into the training of physics-informed neural networks and consistently improve their predictive accuracy by a factor of 50-100x across a range of problems in computational physics.

Despite recent progress, we have to admit that we are still at the very early stages of rigorously understanding the capabilities and limitations of physics-informed neural networks. To bridge this gap we need to answer a series of open questions: What is the relation between stiffness in a given  PDE and stiffness in the gradient flow dynamics of the associated PINN model? Can stiffness in the gradient flow dynamics be reduced (for e.g. using domain decomposition techniques, different choices of loss functions, more effective neural architectures, etc.)? If stiffness turns out to be an inherent property of PINNs, what else can we do to enhance the robustness of their training and the accuracy of their predictions? Can we devise more stable and effective optimization algorithms to train PINN models with stiff gradient flow dynamics? How does stiffness affect the approximation error and generalization error of PINNs? Addressing these challenges calls for a fruitful synergy between deep learning, optimization, numerical analysis and dynamical systems theory that has the potential to crystallize new exciting developments in computational science and engineering.

\section*{Acknowledgements}
This work received support from the US Department of Energy under the Advanced Scientific Computing Research program (grant DE-SC0019116) and the Defense Advanced Research Projects Agency under the Physics of Artificial Intelligence program (grant HR00111890034).

\bibliography{main}
\bibliographystyle{unsrt}

\section{Appendix}
\label{sec:appendix}

\subsection{A bound for the gradients of PINNs boundary and residual loss functions for a one-dimensional Poisson problem}
\label{sec:appendix_proof}

\begin{proof}
Recall that the loss function is given by
\begin{align}
    \begin{split}
         \mathcal{L}(\theta) &=   \mathcal{L}_r(\theta) +  \mathcal{L}_{u_b}(\theta) \\
             &= \frac{1}{N_b} \sum_{i=1}^{N_b}[f_{\theta}(x_b^i) - h(x_b^i)]^2 + \frac{1}{N_r} \sum_{i=1}^{N_r}[\frac{\partial^2}{\partial x^2} f_{\theta}(x_r^i) - g(x_r^i)]^2.
    \end{split}
\end{align}

Here we fix $\theta \in \Theta$, where $\Theta$ denote all weights in a neural network. Then by assumptions, $\frac{\partial \mathcal{L}_{u_{b}}(\theta)}{\partial \theta}$ can be computed by 
\begin{align*}
    \left|\frac{\partial \mathcal{L}_{u_{b}}(\theta)}{\partial \theta}\right| &=\left|\frac{\partial}{\partial \theta}\left(\frac{1}{2} \sum_{i=1}^{2}\left(u_{\theta}\left(x_{b}^{i}\right)-h\left(x_{b}^{i}\right)\right)^{2}\right)\right| \\ &=\left|\sum_{i=1}^{2}\left(u_{\theta}\left(x_{b}^{i}\right)-h\left(x_{b}^{i}\right)\right) \frac{\partial u_{\theta}\left(x_{b}^{i}\right)}{\partial \theta}\right| \\ 
    &=\left|\sum_{i=1}^{2}\left(u\left(x_{b}^{i}\right) \cdot \epsilon_{\theta}\left(x_{b}^{i}\right)-u\left(x_{b}^{i}\right)\right) u\left(x_{b}^{i}\right) \frac{\partial \epsilon_{\theta}\left(x_{b}^{i}\right)}{\partial \theta}\right| \\ &=\left|\sum_{i=1}^{2} u\left(x_{b}^{i}\right)\left(1-\epsilon_{\theta}\left(x_{b}^{i}\right)\right) u\left(x_{b}^{i}\right) \frac{\partial \epsilon_{\theta}\left(x_{b}^{i}\right)}{\partial \theta}\right| \\ 
    & \leqslant\left\|\frac{\partial \epsilon_{\theta}(x)}{\partial \theta}\right\|_{L^{\infty}} \cdot 2 \epsilon 
\end{align*}

Next, we may rewrite the $\mathcal{L}_r$ as 
\begin{align*}
     \mathcal{L}_{r} &=\frac{1}{N_{f}} \sum_{i=1}^{N_{f}}\left|\frac{\partial^{2} u_{\theta}}{\partial x^{2}}\left(x_{f}^{i}\right)-\frac{\partial^{2} u}{\partial x^{2}}\left(x_{f}^{i}\right)\right|^{2}  \approx \int_{0}^{1}\left(\frac{\partial^{2} u_{\theta}(x)}{\partial x^{2}}-\frac{\partial^{2} u(x)}{\partial x^{2}}\right)^{2} d x 
\end{align*}
Then by integration by parts we have,
\begin{align*}
     \frac{\partial \mathcal{L}_{r}}{\partial \theta} 
     &=\frac{\partial}{\partial \theta} \int_{0}^{1}\left(\frac{\partial^{2} u_{\theta}(x)}{\partial x^{2}}-\frac{\partial^{2} u(x)}{\partial x^{2}}\right)^{2} d x \\ 
     &=\int_{0}^{1} \frac{\partial}{\partial \theta}\left(\frac{\partial^{2} u_{\theta}(x)}{\partial x^{2}}-\frac{\partial^{2} u(x)}{\partial x^{2}}\right)^{2} d x \\ 
     &=\int_{0}^{1} 2\left(\frac{\partial^{2} u_{\theta}(x)}{\partial x^{2}}-\frac{\partial^{2} u(x)}{\partial x^{2}}\right) \frac{\partial}{\partial \theta}\left(\frac{\partial^{2} u_{\theta}(x)}{\partial x^{2}}\right) d x \\ &\left.=2 \int_{0}^{1}\left(\frac{\partial^{2} u_{\theta}(x)}{\partial x^{2}}-\frac{\partial^{2} u(x)}{\partial x^{2}}\right) \frac{\partial^{2}}{\partial x^{2}} \frac{\partial\left(u_{\theta}(x)\right.}{\partial \theta}\right) d x \\
    &= 2\left[\frac{\partial^2 u_{\theta}(x)}{ \partial x \partial \theta} \left(\frac{\partial^2 u_{\theta}(x)}{\partial x^2} -  \frac{\partial^2 u(x)}{\partial x^2} \right)\Big|_0^1  - \int_0^1 \frac{\partial^2 u_{\theta}(x)}{ \partial x \partial \theta} \left(\frac{\partial^3 u_{\theta}(x)}{\partial x^3} - \frac{\partial^3 u(x)}{\partial x^3} \right)dx \right] \\
    &=  2 \bigg[\frac{\partial^2 u_{\theta}(x)}{ \partial x \partial \theta} \left(\frac{\partial^2 u_{\theta}(x)}{\partial x^2} -  \frac{\partial^2 u(x)}{\partial x^2} \right)\Big|_0^1 -  \frac{\partial u_{\theta}(x)}{ \partial \theta} \left(\frac{\partial^3  u_{\theta}(x)}{\partial x^3} -  \frac{\partial^3 u(x)}{\partial x^3}\right)\Big|_0^1  \\ 
    &+  \int_0^1 \frac{\partial u_{\theta}(x)}{\partial \theta} \left(\frac{\partial^4 u_{\theta}(x)}{\partial x^4} - \frac{\partial^4 u(x)}{\partial x^4} \right)dx \bigg] \
\end{align*}

Note that 
\begin{align*}
\left|\frac{\partial^{2} u_{\theta}(x)}{\partial x \partial \theta}\right| &=\left|\frac{\partial^{2} u(x) \epsilon_{\theta}(x)}{\partial x \partial \theta}\right| =\left|\frac{\partial}{\partial \theta}\left(u^{\prime}(x) \epsilon_{\theta}(x)+u(x) \epsilon_{\theta}^{\prime}(x)\right)\right| \\ &=\left|u^{\prime}(x) \frac{\partial \epsilon_{\theta}(x)}{\partial \theta}+u(x) \frac{\partial \epsilon_{\theta}^{\prime}(x)}{\partial \theta}\right| 
\leq C\left\|\frac{\partial \epsilon_{\theta}(x)}{\partial \theta}\right\|_{L^{\infty}}+\left\|\frac{\partial \epsilon_{\theta}^{\prime}(x)}{\partial \theta}\right\|_{L^{\infty}}
\end{align*}
And 
\begin{align*}
\left|\frac{\partial^{2} u_{\theta}(x)}{\partial x^{2}}-\frac{\partial^{2} u(x)}{\partial x^{2}}\right| 
&=\left|\frac{\partial^{2} u(x) \epsilon_{\theta}(x)}{\partial x^{2}}-\frac{\partial^{2} u(x)}{\partial x^{2}}\right| \\ 
&=\left|u^{\prime \prime}(x) \epsilon_{\theta}(x)+2 u(x) \epsilon^{\prime \prime}_{\theta}(x)-u^{\prime \prime}(x)\right| \\ &=\left|u^{\prime \prime}(x)\left(\epsilon_{\theta}(x)-1\right)+2 u(x) \epsilon_{\theta}^{\prime \prime}(x)\right| \\ & \leq C^{2} \epsilon+2 \epsilon 
\end{align*}
So we have 
\begin{align}
\left|\frac{\partial^{2} u_{\theta}(x)}{\partial x \partial \theta}\left(\frac{\partial^{2} u_{\theta}(x)}{\partial x^{2}}-\frac{\partial^{2} u(x)}{\partial x^{2}}\right)\Big|_{0}^{1} \right| & \leq 2\left(C\left\|\frac{\partial \epsilon_{\theta}(x)}{\partial \theta}\right\|_{L^{\infty}}+\left\|\frac{\partial \epsilon_{\theta}^{\prime}(x)}{\partial \theta}\right\|_{L^{\infty}}\right)\left(C^{2} \epsilon+2 \epsilon\right) \\ &=O\left(C^{3}\right) \cdot \epsilon \cdot\left\|\frac{\partial \epsilon_{\theta}(x)}{\partial \theta}\right\|_{L^{\infty}} 
\end{align}
Similarly,
\begin{align}
\left|\frac{\partial u_{\theta}(x)}{\partial \theta}\left(\frac{\partial^{3} u_{\theta}(x)}{\partial x^{3}}-\frac{\partial^{3} u(x)}{\partial x^{3}}\right)\Big|_{0}^{1} \right| 
&=\left|\frac{\partial u_{\theta}(x)}{\partial \theta}\left(\frac{\partial^{3} u(x) \epsilon_{\theta}(x)}{\partial x^{3}}-\frac{\partial^{3} u(x)}{\partial x^{3}}\right)\Big|_{0}^{1} \right| \\ 
& \leq O\left(C^{3}\right) \cdot \epsilon \cdot\left\|\frac{\partial \epsilon_{\theta}(x)}{\partial \theta}\right\|_{L^{\infty}}
\end{align}
Finally,
\begin{align}
\left|\int_{0}^{1} \frac{\partial u_{\theta}(x)}{\partial \theta}\left(\frac{\partial^{4} u_{\theta}(x)}{\partial x^{4}}-\frac{\partial^{4} u(x)}{\partial x^{4}}\right) d x\right| \leq O\left(C^{4}\right) \cdot \epsilon \cdot\left\|\frac{\partial \epsilon_{\theta}(x)}{\partial \theta}\right\|_{L^{\infty}}
\end{align}
Therefore, plugging all these together we obtain
\begin{align}
\left|\frac{\partial \mathcal{L}_{r}}{\partial \theta}\right| \leq O\left(C^{4}\right) \cdot \epsilon \cdot\left\|\frac{\partial \epsilon_{\theta}(x)}{\partial \theta}\right\|_{L^{\infty}}
\end{align}

\end{proof}

\subsection{Reference solution for a flow in a two-dimensional lid-driven cavity via a finite difference approximation}
\label{sec:appendix_FDM}
If we introduce the stream function $\psi$ and vorticity $\omega$, the Navier-Stokes equation can be written in the following form \cite{ghia1982high}:
	\begin{equation}
		\left\{
		\begin{aligned}
			&\frac{\partial \omega}{\partial t}+\frac{\partial \psi}{\partial y}\frac{\partial \omega}{\partial x}-\frac{\partial \psi}{\partial x}\frac{\partial \omega}{\partial y}=\nu\Delta \omega\\
			&\Delta \psi=-\omega
		\end{aligned}		
		\right.
		\label{psi_omega}
	\end{equation}
	where
	\begin{equation}
		u=\frac{\partial \psi}{\partial y},\ 
		v=-\frac{\partial \psi}{\partial x},\ 
		\omega=\frac{\partial v}{\partial x}-\frac{\partial u}{\partial y}
		\label{uv_psiomega}
	\end{equation}

	The setup of the simulation is as follows. A set of points $(x_i, y_j)$ is uniformly distributed in the domain $[0,1]\times[0,1]$, with $x_i=i/N$, $y_i=j/N$, $i,j=0,1,\cdots,N$. The grid resolution $h$ equals $1/N$. We denote $A_{i,j}$ as the value of physical variable $A$ (velocity, pressure, etc.) at the point $(x_i,y_j)$. All the spacial derivatives are treated with 2nd-order discretization scheme shown in Eq.\ref{discretization}.
	\begin{equation}
		\begin{aligned}
		\frac{\partial A}{\partial x}|_{i,j}&\approx\frac{A_{i+1,j}-A_{i-1,j}}{2h}\\
		\frac{\partial A}{\partial y}|_{i,j}&\approx\frac{A_{i,j+1}-A_{i,j-1}}{2h}\\
		\Delta A|_{i,j}&\approx\frac{A_{i+1,j}+A_{i-1,j}+A_{i,j+1}+A_{i,j-1}-4A_{i,j}}{h^2}
		\end{aligned}
		\label{discretization}
	\end{equation}

	According to the boundary condition of velocity
	\begin{equation}
		(u,v)=\left\{
		\begin{aligned}
			&(1,0),\ y=1\\
			&(0,0),\ otherwise
		\end{aligned}
		\right.
	\end{equation}
	we can derive the boundary condition of stream function
	\begin{equation}
		\psi=\left\{
		\begin{aligned}
			&h/2,\ &&y=1\\
			&0,\ &&\text{otherwise}\\
		\end{aligned}
		\right.
		\label{psiBC}
	\end{equation}
	under the assumption that the x-component of velocity $u$ grows linearly from $0$ to $1$ between $(0,1-1/N)$ and $(0,1)$, and between $(1,1-1/N)$ and $(1,1)$.

	The boundary condition of vorticity is derived from the Wood's formula \cite{woods1954note}
	\begin{equation}
		\omega_0=-\frac{1}{2}\omega_1-\frac{3}{h^2}(\psi_1-\psi_0)-\frac{3}{h}v_\tau-\frac{3}{2}\frac{\partial v_n}{\partial\tau}+\frac{h}{2}\frac{\partial^2 v_\tau}{\partial \tau^2}
		\label{omegaBC}
	\end{equation}
	where $(\psi_0,\omega_0)$ is the local stream function and vorticity at a boundary point, $(\psi_1,\omega_1)$ is the stream function and vorticity at the adjacent point along the normal direction, $(v_n, v_\tau)$ is the normal and tangential component of velocity, and $\tau$ is the tangential direction.

	The algorithm is composed of the following steps:
	\begin{enumerate}
		\item[Step.1] Set the stream function $\psi$ and vorticity $\omega$ at inner points to zero, and calculate the $\psi$ and $\omega$ at the boundary using equation \ref{psiBC} and Eq.\ref{omegaBC}. Set time $t=0$;
		\item[Step.2] Calculate the vorticity $\omega$ at inner points at the time $t+\Delta t$ with equation \ref{psi_omega}(1), substituting $\partial \omega/\partial t$ with $(\omega(t+\Delta t)-\omega(t))/\Delta t$;
		\item [Step.3] Calculate the stream function $\psi$ at inner points at the time $t+\Delta t$ with equation \ref{psi_omega}(2) and boundary condition equation \ref{psiBC};
		\item[Step.4] Calculate the vorticity $\omega$ on the boundary with equation \ref{omegaBC};
		\item[Step.5] Update the velocity $(u,v)$ at the time $t+\Delta$ with equation \ref{uv_psiomega}, and calculate the error between the velocity at the time $t$ and $t+\Delta t$
		\begin{equation}
			\begin{aligned}
			error_u=\frac{\max\limits_{i,j}\{u_{i,j}(t+\Delta t)-u_{i,j}(t)\}}{\Delta t}\\
			error_v=\frac{\max\limits_{i,j}\{v_{i,j}(t+\Delta t)-v_{i,j}(t)\}}{\Delta t}
			\end{aligned}
		\end{equation}
		If $\max\{error_u,error_v\}<\varepsilon$, the flow has reached the steady state, and the computation terminates. Otherwise, set $t\leftarrow t+\Delta t$, and return to Step.2.
	\end{enumerate}
	
	In our simulation, we set the number of grid $N=128$, time step $\delta t=1\times10^3$, and criterion for convergence $\varepsilon=1\times10^{-4}$.

\end{document}